\documentclass[letterpaper]{article} % DO NOT CHANGE THIS
\usepackage{aaai25}  % DO NOT CHANGE THIS [submission]{aaai25}
\usepackage{times}  % DO NOT CHANGE THIS
\usepackage{helvet}  % DO NOT CHANGE THIS
\usepackage{courier}  % DO NOT CHANGE THIS
\usepackage{fix-cm} % 支持更广泛的 Computer Modern 字体大小
\usepackage[hyphens]{url}  % DO NOT CHANGE THIS
\usepackage{graphicx} % DO NOT CHANGE THIS
\def\UrlFont{\rm}  % DO NOT CHANGE THIS
\usepackage{natbib}  % DO NOT CHANGE THIS AND DO NOT ADD ANY OPTIONS TO IT
\usepackage{caption} % DO NOT CHANGE THIS AND DO NOT ADD ANY OPTIONS TO IT
\usepackage{algorithm}
\usepackage{algorithmic}
\usepackage{subcaption}
\usepackage{enumitem}
\usepackage{calc}
\usepackage{bm}
\usepackage{amsthm}
\usepackage{multirow}
\usepackage{multicol}
\usepackage{pifont}
\usepackage{bbding}
\usepackage{booktabs}  
\usepackage{threeparttable}  
\usepackage{makecell}
\usepackage{graphicx}
\usepackage{subfig}
\usepackage{epstopdf}
\usepackage{float}
\usepackage{fancyhdr}
\usepackage{natbib}
\usepackage{amsfonts} 
\newtheorem{theorem}{Theorem}
\newtheorem{definition}{Definition}
\usepackage{bbding}
\usepackage{amsmath,amssymb}

\usepackage{newfloat}
\usepackage{listings}
\DeclareCaptionStyle{ruled}{labelfont=normalfont,labelsep=colon,strut=off} % DO NOT CHANGE THIS
\floatstyle{ruled}
\newfloat{listing}{tb}{lst}{}
\floatname{listing}{Listing}
\title{Content-aware Balanced Spectrum Encoding in Masked Modeling for Time Series Classification}
\author{
Yudong Han\textsuperscript{\rm 1,\rm 2}\equalcontrib, Haocong Wang\textsuperscript{\rm 1}\equalcontrib, Yupeng Hu\textsuperscript{\rm 1}\thanks{Corresponding author}, Yongshun Gong\textsuperscript{\rm 1}, Xuemeng Song\textsuperscript{\rm 3}, Weili Guan\textsuperscript{\rm 4}
}
\affiliations{
    \textsuperscript{\rm 1}School of Software, Shandong University,
    \textsuperscript{\rm 2}Beijing Institute of Technology, \\
    \textsuperscript{\rm 3}School of Computer Science and Technology, Shandong University,
    \textsuperscript{\rm 4}Harbin Institute of Technology (Shenzhen)
    \{hanyudong.sdu, sduwhc, sxmustc, honeyguan\}@gmail.com, \{huyupeng, ysgong\}@sdu.edu.cn
}

\usepackage{bibentry}
\begin{document}

\maketitle

\begin{abstract}
Due to the superior ability of global dependency, transformer and its variants have become the primary choice in Masked Time-series Modeling (MTM) towards time-series classification task. In this paper, we experimentally analyze that existing transformer-based MTM methods encounter with two under-explored issues when dealing with time series data: (1) they encode features by performing long-dependency ensemble averaging, which easily results in rank collapse and feature homogenization as the layer goes deeper; (2) they exhibit distinct priorities in fitting different frequency components contained in the time-series, inevitably leading to spectrum energy imbalance of encoded feature. To tackle these issues, we propose an auxiliary content-aware balanced decoder (CBD) to optimize the encoding quality in the spectrum space within masked modeling scheme. Specifically, the CBD iterates on a series of fundamental blocks, and thanks to two tailored units, each block could progressively refine the masked representation via adjusting the interaction pattern based on local content variations of time-series and learning to recalibrate the energy distribution across different frequency components. Moreover, a dual-constraint loss is devised to enhance the mutual optimization of vanilla decoder and our CBD. Extensive experimental results on ten time-series classification datasets show that our method nearly surpasses a bunch of baselines. Meanwhile, a series of explanatory results are showcased to sufficiently demystify the behaviors of our method.
\end{abstract}

% Uncomment the following to link to your code, datasets, an extended version or similar.
%
% \begin{links}
%     \link{Code}{https://aaai.org/example/code}
%     \link{Datasets}{https://aaai.org/example/datasets}
%     \link{Extended version}{https://aaai.org/example/extended-version}
% \end{links}

\section{Introduction}
\label{Introduction}
Time-series representation learning has emerged as a fundamental and preposed task for time-series analysis. Different from supervised learning~\cite{hyd_super, DBLP:conf/mm/HanGYLHN21, DBLP:journals/tip/HuLSGN21, DBLP:journals/tip/HuNLWWH21, DBLP:journals/tois/HuWLTN24}, which heavily relies on human-labeled ground truth, self-supervised learning~\cite{nie2022time, tonekaboni2021unsupervised, ozyurt2022contrastive} has shown its flexibility using the intrinsic characteristic of data itself towards scalable representation learning. The widely embraced scheme that initially introduced from computer vision domain~\cite{he2022masked}, Masked Time-series Modeling (MTM), randomly masks a portion of input timestamps, and then reconstructs the invisible timestamps based on the visible ones. Recently, transformer and its variants have become the predominant choice in MTM, and achieve new state-of-the-art performance on various classification benchmarks~\cite{dong2023simmtm, cheng2023timemae}. This greatly attributes to its powerful modeling ability of long-range dependency between different timestamps, which facilitates learning the context-aware feature.
% a series of studies~\cite{dong2023simmtm, cheng2023timemae} have been devoted to further optimizing the masked representation learning.

\begin{figure}[t]
	\centering
	{\includegraphics[width=0.96\linewidth]{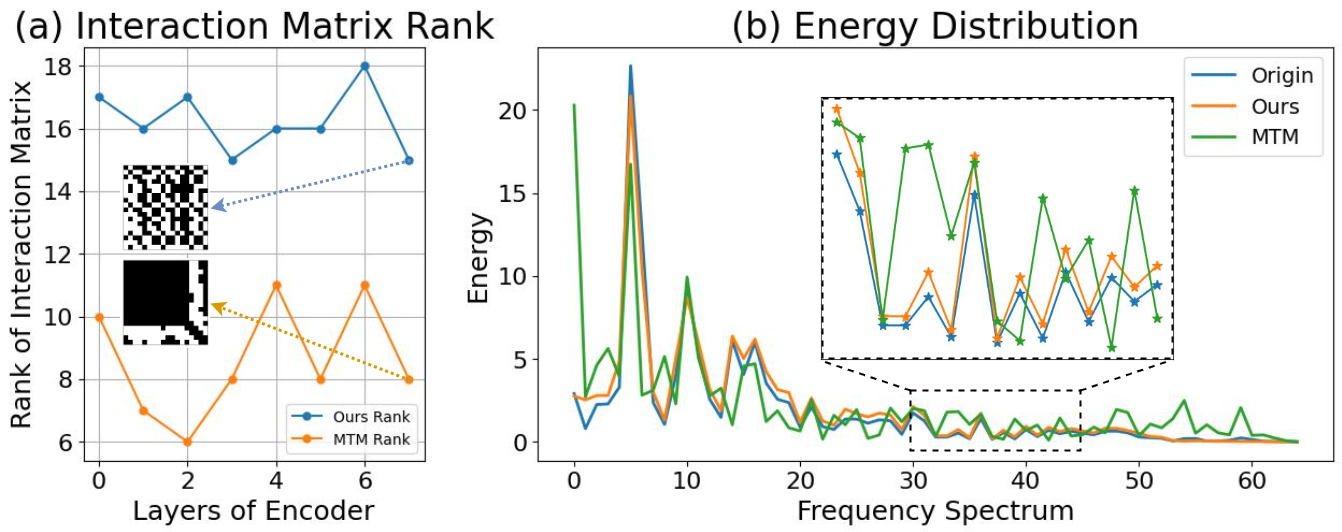}}	
	\caption{(a) Comparison of rank of the interaction matrix across different layers of the encoder in vanilla MTM and our method; (b) Energy distribution comparison of raw data, reconstructed results of vanilla MTM, and that of our method.} 
	\label{fig:intro}
\end{figure}

Despite their promising performance towards representation learning, these transformer-based methods overlook two potential problems: \textbf{(1) feature homogenization.} Several studies~\cite{DBLP:conf/icml/DongCL21, DBLP:conf/iccv/HanPHSH23, DBLP:conf/aaai/HanHSTXN24} point out that the feature encoded by transformer-based backbone easily incurs rank collapse due to the long-dependency \textit{ensemble averaging}~\cite{DBLP:conf/iclr/ParkK22}, and we experimentally conclude that this phenomenon also exists when encoding the time-series data. 
% more quicker and obvious when handling the low-rank (i.e., has more similar sub-timestamps) input series with periodicity and seasonality~\cite{DBLP:conf/icml/DongCL21}. 
As illustrated in Figure \ref{fig:intro} (a), we showcase the learned interaction matrix and calculate their ranks from the encoder of vanilla MTM~\cite{vaswani2017attention} and our improved method. We observe that vanilla MTM tends to produce low-rank features, which manifests to encode homogenized information and lacks sufficient semantic richness.
\textbf{(2) energy imbalance.} Based on \textit{frequency principle}~\cite{Xu_2020, DBLP:conf/icml/RahamanBADLHBC19} and architecture-induced frequency preference~\cite{DBLP:conf/iclr/ParkK22}, we observe that vanilla transformer-based feature learning is inclined to capture low-frequency energy of time-series. In other words, they will easily memorize the sketchy trend of time-series but need more steps to comprehend variation details conveyed by mid/high-frequency energy. As depicted in Figure \ref{fig:intro} (b), we transform the original time-series and reconstructed ones into the Fourier domain respectively, and show their difference of spectrum energy distribution. It can be seen that the reconstruction result from vanilla MTM primarily concentrates on low-frequency energy and lacks adequate attention to other frequency band information, thereby leading to the obvious inconsistency with original time-series. 

However, how to tackle above issues within MTM remains an under-explored problem. In this paper, we theoretically analyze that (1) the rank of encoding feature intrinsically hinges on property of its corresponding spectrum space, and (2) the distribution of feature space could also be directly revealed by the energy distribution in the spectrum space. Based on above considerations, we reveal the possibility of resorting to the unified spectrum encoding to address them. Specifically, we propose an auxiliary content-aware balanced spectrum decoding branch to achieve the spectrum reconstruction in MTM, where several cascaded blocks with two tailored units are employed to motivate the encoder to retard homogenized and imbalanced feature learning, respectively. \textbf{Content-aware Interaction Modulation (CIM) Unit} is first employed, which maps the intermediate representations from encoder into spectrum space and modulates them with a content-aware complex network~\cite{DBLP:journals/corr/TrabelsiBSSSMRB17}. According to $\textit{frequency-domain convolutional theory}$~\cite{DBLP:conf/iccv/00140LZLG23}, this process essentially equals to dynamic kernel convolutional operation, which not only restricts unnecessary interaction scope compared with long-dependency tactic but also has the flexibility to adjust the receptive field based on variation of content. Following CIM, \textbf{Spectrum Energy Rebalance (SER) Unit}, inspired by Bernstein approximation, is leveraged to adjust the energy distribution of encoded feature across different frequency components, thus compensating for overlooked details in mid/high frequency bands. Moreover, dual-constraint loss is further devised to reduce the gap between temporal space and spectrum space, thereby avoiding the information loss in two decoding branches. 

We conduct experiments on ten classification datasets from diverse domains with both univariate and multivariate settings, to demonstrate the superiority of our method in representation learning. More importantly, the related visualization and mechanism exploration results sufficiently indicate the behavior of our model.

In summary, our main contributions are as follows:
\begin{itemize}
    \item We propose a two-pronged masked time-series reconstruction framework, which seamlessly integrates the content-aware balanced decoder into vanilla temporal decoder, along with devised dual-constraint loss, to establish the connection between temporal domain and spectrum domain.
    \item Towards content-aware balanced decoder, we endow it with two iterative units, to effectively retard feature homogenization and achieve spectrum energy rebalance.
    \item We conduct extensive experiments on numerous real-world datasets to validate the effectiveness of our proposed CBD, the superiority of our overall model compared with state-of-art methods, and convincingly reveal the intriguing behavior of our model.
\end{itemize}

\begin{figure*}
 \centering
 \includegraphics[width=170mm]{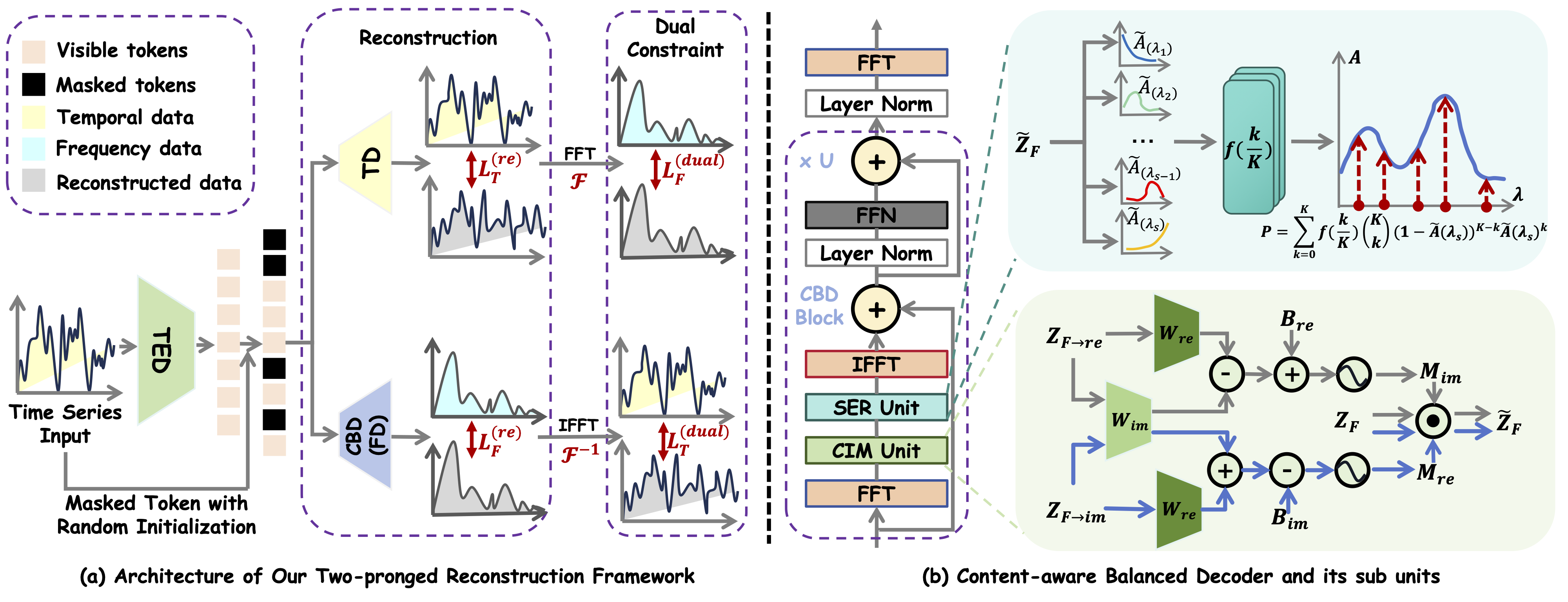}
 \vspace{-1em}
 \caption{(a) Schematic illustration of our two-pronged framework. TED denotes temporal encoder, and TD and CBD represents temporal decoder and frequency decoder (content-aware balanced decoder). (b) The details of content-aware balanced decoder.}
 \label{fig:framework}
\end{figure*}
 \vspace{-1em}

\section{Related Work}

\paragraph{Time-Series Pretraining}
Recent self-supervised Time-Series pre-training techniques can be roughly categorized into two groups: \textit{Time-Series Contrastive Learning} and \textit{Masked Times-Series Modeling}. 

\textbf{Time-Series Contrastive Learning} constructs multiple views to increase variance and align features for robust global representation learning. Existing work~\cite{zhang2022self} focus on designing different views for contrast to capture different preference on learned representation. For example, TS2Vec~\cite{yue2022ts2vec} utilizes temporal contrast to improve discrimination of dynamic semantic variations. TS-TCC~\cite{eldele2021time} formulates a cross-view prediction task that incorporates both temporal and contextual contrast. TimesURL~\cite{DBLP:conf/aaai/LiuC24} enhances learning process through frequency-temporal augmentation and double universum construction. More flexibly, TS-GAC~\cite{wang2024graphaware} emphasizes the role of spatial consistency in a graph contrastive framework. CSL~\cite{DBLP:journals/corr/abs-2305-18888} leverages shapelet-based learning to better fit series data.

Concurrent with these strands of research, \textbf{Masked Time-Series Modeling} learns representations by reconstructing masked timestamps from partial observations. TST~\cite{zerveas2021transformer} marks the first attempt to employ a transformer-based encoder for this task. PatchTST\cite{DBLP:conf/iclr/NieNSK23} predicts masked subseries-level patches to capture local semantics. SimMTM~\cite{dong2023simmtm} argues that random masking would disrupt temporal variations, thereby proposing to aggregate the point-wise representations derived from multiple masked variations for robust reconstruction. To mitigate the gap between masked representation in pre-trained stage and unmasked ones in fine-tuning stage, TimeMAE~\cite{cheng2023timemae} leverages a decoupled scheme to respectively encode the semantic-enhanced representation with high information density. In this paper, we primarily focus on optimizing the masked time-series modeling. Different from above studies, our method reveals two crucial issues in existing transformer-based masked modeling framework: \textit{feature homogenization} and \textit{energy imbalance} from the spectrum perspective.

\paragraph{Learning from Spectrum Perspective}
More and more investigations delve into learning the robust representation by transforming the temporal domain to frequency domain due to its exclusive properties, which have been actively applied to many fields, such as graph classification, domain generalization, and time-series. For example, FACT~\cite{xu2021fourier} reckons that model that highlights phase information could better deal with cross-domain setting, thus developing a Fourier-based data augmentation strategy.  SFA~\cite{zhang2023spectral} introduces a spectrum feature alignment method to address the issue of imbalanced feature when performing graph contrast. In the field of time-series analysis, FEDformer~\cite{DBLP:conf/icml/ZhouMWW0022} launches a frequency-enhanced transformer to emphasize the global view of learned feature for time series forecasting. However, how to effectively retard homogenized and imbalanced feature learning from the spectrum perspective in masked time-series framework still remains an under-explored challenge. 

\section{Methodology}
\label{section2}
% In this section, we first review the masked modeling framework towards time-series data. 
% % Then, the key parts of our framework and loss function are elaborated: Content-aware Balanced Decoder (CBD).
% Then, we conduct an in-depth analysis to the key components of our framework, Content-aware Balanced Decoder, and elaborate on the implementation of its two sub-units.
% Finally, we provide the design details of loss function that optimizes the overall network.

In this section, we begin by reviewing the masked modeling framework towards time-series data. Following this, we delve into an in-depth analysis to the key components of our framework, Content-aware Balanced Decoder (CBD), and elaborate on the implementation of its two tailored units. Finally, we provide the design details of loss function that optimizes the overall network.

\subsection{Review of Masked Time-Series Modeling}
Masked Time-Series Modeling (MTM) adopts a transformer-based architecture to reconstruct large portions of masked timestamps via the visible timestamps. Formally, given $\{x_i\}_{i=1}^N$ as a batch of $N$ time-series samples, where $x_i\in \mathbb{R}^{L\times C}$ contains $L$ timestamps and $C$ variables, we randomly mask a certain proportion of timestamps along the temporal dimension. Following the orthodox pipeline of masked modeling framework, MTM encoder takes these visible timestamps as input to generate an intermediate representation $\mathbf{Z}=\{\mathbf{z}_i\}_{i=1}^N$, while the following decoder achieves temporal reconstruction of masked timestamps. By aligning the reconstruction results with unmasked timestamps, the encoder could learn the representation that capture context-aware dependencies within time-series data.

\subsection{Content-aware Balanced Decoder}
As suggested by MAE~\cite{he2022masked}, the decoder design is crucial to the MTM model, as it not only models the relationship between representations of masked tokens and visible tokens, but also determines the semantic preference of learned representation. To address the issues of feature homogenization and spectrum energy imbalance in traditional transformer-based MTM methods, we armed the temporal encoder (TD) with an auxiliary content-aware balanced decoder (CBD), where two tailored units are iteratively deployed. The overall architecture and unfold details are illustrated in Figure \ref{fig:framework}. In what follows, we elaborate these two units sequentially.

% Specifically, it consists of two parts: Two-Pronged Branch and Dual Constraint. In the two-pronged branch, we deploy an additional frequency decoder (Rectified Spectrum Perceiver), which not only reconstructs frequency information but, more importantly, addressing existing issues by Content-aware Scope Modulation and spectrum energy rebalance. Moreover, we further design the Dual Constraints to ensure that each branch can still retain the information from the other branch: (1) The restored spectrum information still retains local fine-grained details;  (2) The reconstructed temporal information could preserve the global semantics that spectrum perspective brings. We hope the information in temporal and frequency space can play as the reciprocal parts on each other, thereby collaboratively learn the robust representation.

% \subsection{Spectrum-aware Adjustment}
% In this part, we follow the divide-and-conquer philosophy, and design two novel targeted units in the basic cell of Rectified Spectrum Perceiver (RSP): Content- aware Scope Modulation Unit (CSM) and Spectrum Energy Rebalance Unit (SER).

\paragraph{Content-aware Interaction Modulation Unit (CIM)} Inspired by the fact that the rank of feature matrix can be determined by its spectrum distribution (i.e., lower bound), $rank(\mathbf{Z}) \geq rank(\bm{\Sigma})$, where $\bm{\Sigma}$ are the eigenvalue of $\mathbf{Z}$, we devise a content-aware modulation strategy that operates directly in the spectral space. Specifically, 
% which not only restricts unnecessary interaction scope compared with long-dependency tactic but also
% has the flexibility to adjust the receptive field based on varia-
% tion of conten
% spectrum information conveys the accurate holistic characteristics of time series, imposing a global semantic constraint on encoded features from a frequency perspective can mitigate the homogenization caused by over-smoothing problem in vanilla MTM. 
we first transform the intermediate feature $\mathbf{Z}$ to discrete Fourier domain,
\begin{equation}
\fontsize{9pt}{9pt}
\mathbf{Z}_{F} = \mathcal{F}(\mathbf{Z}),
\end{equation}
where $\mathcal{F}$ denotes the Fourier transformation, and each of components $\mathbf{Z}_{F} $ can be written as,
\begin{equation}
\fontsize{9pt}{9pt}
 \mathbf{Z}_{F}(\lambda_{s}) = \sum_{t=1}^{T}\mathbf{Z}(t) \cdot e^{-j \frac{ 2\pi \lambda_{s} t}{T}},
\end{equation}
and the inverse Fourier transformation $\mathcal{F}^{-1}$ corresponds to 
\begin{equation}
\fontsize{9pt}{9pt}
\mathbf{Z}(t) = \frac{1}{T} \sum_{t=1}^{T} \mathbf{Z}_{F}(\lambda_{s})\cdot e^{-j \frac{2\pi \lambda_{s} t}{T}},
\end{equation}
where $T$ denotes the total length of intermediate feature of time-series, and $\lambda_{s}$ denotes the $s$-th frequency component in frequency domain. The real and imaginary part of $\mathbf{Z}_{F}$ can be denoted as $\mathbf{Z}_{F \rightarrow re}$ and $\mathbf{Z}_{F \rightarrow im}$, which exhibits equal contribution for frequency modeling~\cite{DBLP:journals/corr/TrabelsiBSSSMRB17}. In order to achieve more exhaustive spectrum modulation, we encode the real and imaginary parts separately to generate the modulation signals, which is summarized as,
\begin{equation}
\fontsize{9pt}{9pt}
\mathbf{M}(\lambda_{s}) = \sigma(\mathbf{W}_{F} \mathbf{Z}_{F}(\lambda_{s}) + \mathbf{b}_{F}),
\end{equation}
where $\sigma$ is the activation function, $\mathbf{W}_{F} = \mathbf{W}_{F \rightarrow re} + j \cdot \mathbf{W}_{F \rightarrow im}$ and $\mathbf{b}_{F} = \mathbf{b}_{F \rightarrow re} + j \cdot \mathbf{b}_{F \rightarrow im}$. More specifically, the generated $\mathbf{M}$ can be further unfolded into real and imaginary parts as follows,

% \begin{equation}
%     % \fontsize{7.5pt}{7.5pt}
% \begin{aligned}
%     \mathbf{M}_{re}(\lambda_{s}) & = \sigma(\mathbf{W}_{F \rightarrow re} \mathbf{Z}_{F \rightarrow re}(\lambda_{s})  \\
%     & - \mathbf{W}_{F \rightarrow im}\mathbf{Z}_{F \rightarrow im}(\lambda_{s}) + \mathbf{B}_{F \rightarrow re}), 
% \end{aligned}
% \end{equation}

% \begin{equation}
% % \fontsize{7.5pt}{7.5pt}
% \begin{aligned}
%     \mathbf{M}_{im}(\lambda_{s}) &= \sigma(\mathbf{W}_{F \rightarrow im} \mathbf{Z}_{F \rightarrow im}(\lambda_{s}) \\
%     & + \mathbf{W}_{F \rightarrow r} \mathbf{Z}_{F \rightarrow re}(\lambda_{s}) + \mathbf{B}_{F \rightarrow im}), 
% \end{aligned}
% \end{equation}

\begin{equation}
\fontsize{9pt}{9pt}
    \left\{ \begin{array}{ll} 
    \begin{aligned}
    \mathbf{M}_{re}(\lambda_{s}) = \sigma & (\mathbf{W}_{F \rightarrow re} \mathbf{Z}_{F \rightarrow re}(\lambda_{s}) \\ 
    & - \mathbf{W}_{F \rightarrow im}\mathbf{Z}_{F \rightarrow im}(\lambda_{s}) + \mathbf{B}_{F \rightarrow re}), \\
    
    \mathbf{M}_{im}(\lambda_{s}) = \sigma & (\mathbf{W}_{F \rightarrow im} \mathbf{Z}_{F \rightarrow im}(\lambda_{s}) \\
    & + \mathbf{W}_{F \rightarrow r} \mathbf{Z}_{F \rightarrow re}(\lambda_{s}) + \mathbf{B}_{F \rightarrow im}). \\
    \end{aligned}
    \end{array} \right.
\end{equation}
Afterwards, the generated complex signal is used to modulate counterparts of original frequency-domain feature, which can be written as,
\begin{equation}
\fontsize{9pt}{9pt}
\mathbf{\tilde{Z}}_{F}(\lambda_{s}) = \mathbf{M}(\lambda_{s}) \odot \mathcal{F}(\mathbf{Z}).
\end{equation}
Based on \textit{theorem} 1, Eq.(6) essentially equals to a dynamic convolutional operation. In a deeper analysis from the perspective of temporal domain, it serves as two purposes: 
(1) Unlike the long-range dependency mechanism in self-attention block, convolutional operation leads to less freedom of interaction, which naturally restore the rank of features~\cite{DBLP:conf/iccv/HanPHSH23}.
(2) its dynamicity indicates that the learned kernel is determined by the content variation of each timestamp, which effectively diversifing the interaction pattern. In other words, for smoothing sub-series, a large interaction scope is possibly unnecessary, while for sub-series that has more obvious fluctuation, enlarging the interaction scope to obtain more context is crucial for understanding its comprehensive semantic. By means of Eq.(6),  feature homogenization and rank collapse can be effectively mitigated via direct spectrum modulation. The proof of \textit{theorem} 1 is illustrated in supplementary materials.     
\begin{theorem}
(Frequency-domain convolution theorem)  The multiplication of two signals in the Fourier domain equals to Fourier transformation of a convolution of these two signals in temporal domain, which can be summarized as,
\begin{equation}
\fontsize{9pt}{9pt}
\mathcal{F}\left [\mathbf{K}(t) \otimes \mathbf{Z}(t) \right] = \mathcal{F}(\mathbf{K}(t)) \odot \mathcal{F}(\mathbf{Z}(t)),
\end{equation}
where $\otimes$ and $\odot$ denote the convolutional operation and element-multiplication operation, respectively, $\mathbf{K}(t)$ and $\mathbf{Z}(t)$ represent two signals with respect with time variable $t$, and $\mathcal{F}(\cdot)$ denotes the Fourier transformation.
\end{theorem}

\paragraph{Spectrum Energy Rebalance Unit (SER)} For time-series data, timestamps with obvious fluctuations generally contain more high-frequency components~\cite{DBLP:conf/iccv/QinZW021} in spectrum space. However, several previous literatures indicate that networks tend to prioritize learning low-frequency components over high-frequency ones\cite{DBLP:conf/aaai/XuZ21,DBLP:conf/iconip/XuZX19}, resulting in challenges in capturing high-frequency details. Furthermore, according to \textit{theorem} 2, overlooking the certain bands of frequency components would greatly impact the temporal information encoding, thereby hindering balanced representation learning. The proof of \textit{theorem} 2 is shown in supplementary materials.

\begin{theorem}
(Parseval’s theorem) Assume that $\mathbf{Z}$ denotes the original temporal domain feature, and $\mathbf{Z}_{F}$ is the corresponding spectrum representation, then the energy of feature in temporal domain equals to the energy in frequency domain,
\begin{equation}
\fontsize{9pt}{9pt}
\sum_{t=-\infty}^{\infty} |\mathbf{Z}(t)|^{2} = \sum_{s=-\infty}^{\infty} |\mathbf{Z}_{F}(\lambda_{s})|^{2}, 
\end{equation}
where $\mathbf{Z}_{F} = \mathcal{F}(\mathbf{Z})$, $\mathcal{F}(\cdot)$ denotes the Fourier transformation, $\lambda_{s}$ represents the $s$-th spectrum component, and $t$ denotes the time variable
\end{theorem}

To tackle it, we design Spectrum Energy Rebalance unit (SER) to flexibly adjust the energy distribution in spectrum space to approximate that of original time-series data. We assume that the ideal response function from spectrum components to energy can be denoted as $\mathcal{G}(\cdot)$, and the initial response is $\mathbf{A}=\sqrt{\mathbf{\tilde{Z}}_{F\rightarrow re}^2+\mathbf{\tilde{Z}}_{F\rightarrow im}^2} = \Vert \mathbf{\tilde{Z}}_{F} \Vert_{2} $. SER serves as the energy rebalance mapping function $\mathcal{P}(\cdot)$, which makes the initial response $\mathbf{A}$ head to real response $\mathcal{G}(\mathbf{\tilde{Z}}_{F})$, i.e., $\mathcal{G}(\mathbf{\tilde{Z}}_{F})=\mathcal{P}(\cdot) \circ \Vert \cdot \Vert_{2} (\mathbf{\tilde{Z}}_{F})$, where $\circ$ denotes the compound operation. To better simulate the effect of $\mathcal{P}(\cdot)$, we adopt Bernstein polynomials. Before unfolding our design details of SER, we first give the definition of \textit{Bernstein polynomial approximation} in the following.

\begin{definition}
(Bernstein polynomial approximation) Given an arbitrary continuous mapping function $f(w)$ from $w \in \left [0, 1 \right ]$, the K-order Bernstein polynomial approximation of $f(w)$ is defined as,
\begin{equation}
\fontsize{9pt}{9pt}
p_{K}(w) = \sum_{k=0}^{K} \Theta_{k} \cdot \mathcal{B}_{k}^{K}(w) = \sum_{k=0}^{K}f(\frac{k}{K}) \cdot \binom{K}{k} (1-w)^{K-k}w^{k},
\end{equation}
and we have $p_{K}(w) \rightarrow f(w)$ as $K \rightarrow \infty$, where $\Theta_{k}$ works as the coefficient of $\mathcal{B}_{k}^{K}(w)$, and $\mathcal{B}_{k}^{K}(w)$ serves as the base of polynomial.
\end{definition}

Given that the domain of definition of $p_{K}(w)$ is restricted to $\left[ 0, 1\right]$, we therefore adopt the softmax function to normalize the input energy distribution $\mathbf{A}(\lambda_{s})$, 
% \begin{equation}
% \fontsize{9pt}{9pt}
% \mathcal{\tilde{A}}(\lambda_{s}) = \frac{e^{\mathcal{A}(\lambda_{s})}}{e^{\mathcal{A}(\lambda_{s})} + \sum_{i \notin s}e^{\mathcal{A}(\lambda_{s})}},
% \end{equation}
and we substitute the normalized energy distribution $\mathbf{\tilde{A}}(\lambda_{s})$ into the approximated polynomial $p_{K}(w)$,
\begin{equation}
\fontsize{9pt}{9pt}
   \begin{aligned}
   p_{K} = \sum_{k=0}^{K} f(\frac{k}{K}) \cdot \binom{K}{k} (1-\mathbf{\tilde{A}}(\lambda_{s}))^{K-k}\mathbf{\tilde{A}}(\lambda_{s})^{k}.
   \end{aligned}
\end{equation} 
Beyond that, in order to adjust $\mathcal{P}$ (i.e., $p_{K}$) in a data-driven manner, we leverage an energy-aware gating network to control the filter coefficients,
\begin{equation}
\fontsize{9pt}{9pt}
f(\frac{k}{K}) = (\mathbf{W}_{c} \mathbf{A}^{\dagger} + \mathbf{b}_{c})_{k},
\end{equation}
where $\mathbf{A}^{\dagger}= \left [\mathbf{\tilde{A}}(\lambda_{1}),\mathbf{\tilde{A}}(\lambda_{2}), ..., \mathbf{\tilde{A}}(\lambda_{S}) \right ] \in \mathbb{R}^{S}$. $\mathbf{W}_{c} \in \mathbb{R}^{K \times S}$ and $\mathbf{b}_{c} \in \mathbb{R}^{K}$ are the learnable weights and bias of gating network, respectively.

Such mechanism potentially undertakes the role that it allows the model to have elasticity to compensate for the missing spectrum components in the training process. Interestingly, from a graph perspective, different ways of frequency response can be interpreted as different specific filters~\cite{DBLP:conf/iconip/XuZX19, DBLP:journals/pami/BianchiGLA22}. When $K$ is sufficiently large, function $\mathcal{P}$ could simulate arbitrary spectrum rebalance functions (i.e., filters). Finally, the learned spectrum rebalance function is utilized to modulate the frequency component to obtain the balanced spectrum feature,

\begin{equation}
\fontsize{9pt}{9pt}
\mathbf{\tilde{Z}_{F\rightarrow{bal}}}(\lambda_{s}) = p_{K}(\lambda_{s}) \odot \mathbf{\tilde{Z}}_{F}(\lambda_{s}).
\end{equation}

\paragraph{Iterative Architecture}
As illustrated in Figure 2(b), it is built upon the vanilla transformer architecture, which modifies the self-attention block by integrating CIM and SER unit. For each layer $u \in \left [1, U \right]$, where $U$ is the total number of layer, a CBD block processes and updates the masked representation as follows,
\begin{equation}
\fontsize{9pt}{9pt}
\mathbf{\tilde{Z}}^{(u)}_{F} = \mathrm{CBDBlock}(\mathbf{\tilde{Z}}^{(u-1)}_{F};\bm{\Omega}^{(u)}),
\end{equation}
where $\bm{\Omega}^{(u)} = \{\mathbf{W}_{F}^{(u)}, \mathbf{b}_{F}^{(u)}, \mathbf{W}^{(u)}_{c}, \mathbf{b}^{(u)}_{c}\}$ denotes the set of learned parameters in each block. At each step, CBD block retards rank collapse of interaction matrix and mitigates feature smoothing via CIM unit, and further leverages SER unit to balance spectrum energy. Moreover, each block has independent parameters, allowing blocks at different depths to collaboratively refine the masked representation.

\subsection{Optimization Strategy}
\paragraph{Dual-Constraint Pre-trained Loss} During the pre-training stage, as shown in Figure 2(b), temporal encoder (TED) encodes the masked time series and decodes it into two outputs: temporal reconstruction $\mathbf{\tilde{Z}}^{(U)}_{T}$ and frequency reconstruction $\mathbf{\tilde{Z}}^{(U)}_{F}$. We denote the ground truth of temporal domain and its frequency domain as $\mathbf{Z}^{(gt)}_{T}$ and $\mathbf{Z}^{(gt)}_{F}$, the pre-training loss is formulated as:

\begin{equation}
\fontsize{9pt}{9pt}
   \begin{aligned}
\mathcal{L} &= \mathcal{L}_{T}^{(re)}(\mathbf{\tilde{Z}}_{T}, \mathbf{Z}^{(gt)}_{T}) + \mathcal{L}_{F}^{(dual)}(\mathcal{F}({\mathbf{\tilde{Z}}_{T}}), \mathbf{Z}^{(gt)}_{F}) \\
& + \gamma \left [\mathcal{L}_{F}^{(re)}(\mathbf{\tilde{Z}}^{(U)}_{F}, \mathbf{Z}^{(gt)}_{F}) + \mathcal{L}_{T}^{(dual)}(\mathcal{F}^{-1}(\mathbf{\tilde{Z}}^{(U)}_{F}), \mathbf{Z}^{(gt)}_{T}) \right],
   \end{aligned}
\end{equation}
where $\mathcal{L}^{(re)}_{\sim}$ represents the reconstruction loss in traditional masked modeling, while $\mathcal{L}^{(dual)}_{\sim}$ serves as the dual-constraint loss aforementioned. For temporal domain loss $\mathcal{L}^{\sim}_{T}$, we compute Mean Square Error (MSE) between the reconstructed and raw time-series. As to frequency domain loss $\mathcal{L}^{\sim}_{F}$, we compute squared Euclidean distance between their real and imaginary parts. $\gamma$ is the loss weight to control the contribution of each branch. Specially, the dual-constraint loss ensures that the two decoding branches can enhance the information consistency between temporal domain and frequency domain in the training process.

\paragraph{Fine-tuned Loss}
In the fine-tuning stage, the output $\mathbf{Z}$ from the common encoder is fed into the final classification head $f_c$. Then the fine-tuned loss is termed as,
\begin{equation}
\fontsize{9pt}{9pt}
    \label{eq:15}
    \mathcal{L} = \mathcal{L}_{CE}(f_c(\mathbf{Z}), y),
\end{equation}
where the Cross-Entropy Loss is utilized to minimize the distance between label $y$ and the prediction $f_c(\mathbf{Z})$.

% In the fine-tuning stage, we discard both the temporal decoder and frequency decoder, and only retain the common encoder for downstream task adaption. The high-semantic representation outputed from encoder $\mathbf{X}_{T}^{e}$ is fed into the classification head, which is implemented by a FC layer,
% \begin{equation}
%     \label{eq:15}
%     \mathbf{h} = f_{d \rightarrow |\mathcal{C}|}(\mathbf{X}_{T}^{e}),
% \end{equation}
% where $\mathcal{C}$ denotes the candidate classes, $|\mathcal{C}|$ represents the number of element in $\mathcal{C}$, $d$ denotes the feature dimension before predictor, and $\mathbf{h} \in \mathbb{R}^{|\mathcal{C}|}$ represents the predicted answer distribution. Following the widely embraced loss for classification task, Cross-Entropy loss function is utilized to minimize the distribution distance between ground-truth and the prediction.

\section{Experiments}
In this section, we first introduce the experiment setup. Then, we present and analyze the quantitative results of our model and a bunch of baselines on time series classification. Finally, we demonstrate the effectiveness of our model through analytical experiments.

\begin{table*}[htbp]
  \centering
  % \begin{tiny}
  \fontsize{7pt}{4pt}\selectfont
  \setlength{\tabcolsep}{4pt}
    % \begin{adjustbox}{width = 1.\textwidth,center}
    % \resizebox{1.0\textwidth}{!}{%
     % \setlength{\tabcolsep}{0.1mm}
    \begin{tabular}{c|cccccccccc}
    \toprule
    \toprule
    \multirow{2}{*}{Methods} & \multicolumn{10}{c}{\textit{Linear Evaluation}} \\
     & HAR & PS & SRSCP1 & MI & FM & AWR & SAD & ECG5000 & FB & UWare \\
    \midrule
    \vspace{2pt}
    \multirow{2}{*} \textit{TS-TCC(IJCAI'21)} & 89.22$\pm$0.70 &   14.27$\pm$0.39 & 83.64$\pm$0.99 & 55.47$\pm$0.68 & 48.00$\pm$1.63 & 93.23$\pm$0.68 & 95.20$\pm$0.15 & 92.67$\pm$0.78 & 50.29$\pm$0.31 & 83.17$\pm$0.29\\
          % & \textit{MF1}   & 89.23$\pm$0.76 &  12.23$\pm$0.35 &  57.83$\pm$2.24 & 38.00$\pm$0.75 &  47.36$\pm$1.64 & 90.94$\pm$0.73 &  95.24$\pm$0.15 & 52.33$\pm$0.58 & 48.03$\pm$0.04 & 82.09$\pm$0.07\\
    \vspace{2pt}
    \multirow{2}[2]{*} \textit{TS2Vec(AAAI'22)} & 90.36$\pm$0.32 &   10.82$\pm$0.38 &  83.61$\pm$0.99 & 51.00$\pm$0.75 & 47.10$\pm$4.22 & 98.30$\pm$0.09 & 97.31$\pm$0.19 & 93.50$\pm$0.65 & 78.90$\pm$0.54 & 93.40$\pm$0.53\\
          % & \textit{MF1}   & 90.36$\pm$0.33 &  10.09$\pm$0.26 & 82.92$\pm$1.68 & 47.31$\pm$0.64 & 47.03$\pm$4.18 &  98.29$\pm$0.10 &  97.31$\pm$0.19 & 50.24$\pm$0.68 & 73.30$\pm$0.28 & \underline{93.35$\pm$0.41}\\
    
    %\multirow{2}[2]{*}{MHCCL} & \textit{Acc.}  & 82.95$\pm$0.55 &  00.00$\pm$0.00 & 55.26$\pm$1.12 & 00.00$\pm$0.00 & 52.40$\pm$2.28 & 93.00$\pm$0.56 & 95.91$\pm$0.56 & 00.00$\pm$0.00 & 00.00$\pm$0.00 & 00.00$\pm$0.00\\
          %& \textit{MF1}   & 82.70$\pm$0.62 &  00.00$\pm$0.00 & 53.10$\pm$1.92 & 00.00$\pm$0.00 & 49.82$\pm$3.06 & 93.14$\pm$0.75 & 95.92$\pm$0.45 & 00.00$\pm$0.00 & 00.00$\pm$0.00 & 00.00$\pm$0.00\\
    %\midrule
    \vspace{2pt}
    \multirow{2}[2]{*} \textit{TS-GAC(AAAI'24)} & \underline{91.40$\pm$0.16} &  13.53$\pm$0.48 & 86.04$\pm$0.31 & 56.00$\pm$0.46 & 53.39$\pm$0.54 & \underline{98.50$\pm$0.06} & \textbf{97.99$\pm$0.05} & - & - & - \\
          % & \textit{MF1}   & 90.70$\pm$0.00 &  12.85$\pm$0.46 & 85.54$\pm$0.51 & 50.25$\pm$0.36 & 49.58$\pm$0.71 & \underline{98.40$\pm$0.07} & \textbf{97.99$\pm$0.05} & - & - & - \\

    \vspace{2pt}
    \multirow{2}[2]{*} \textit{TimesURL(AAAI'24)} & 87.18$\pm$0.46 & 16.88$\pm$0.42 & 85.32$\pm$0.95 & 61.20$\pm$0.35 & 56.00$\pm$1.84 & 98.00$\pm$0.13 & 95.86$\pm$0.10 & 93.73$\pm$0.58 & 73.95$\pm$0.34 & 94.22$\pm$0.38 \\

    \vspace{2pt}
    \multirow{2}[2]{*} \textit{CSL(VLDB'24)} & 83.40$\pm$0.33 & \underline{17.63$\pm$0.36} & 84.60$\pm$0.74 & 61.00$\pm$0.45 & \underline{56.25$\pm$2.63} & 97.67$\pm$1.35 & 94.95$\pm$0.26 & 93.16$\pm$0.84 & \underline{79.50$\pm$0.42} & \underline{95.31$\pm$0.20} \\
    
    \midrule
    %\multirow{2}[2]{*}{TST} & \textit{Acc.}  & 87.39$\pm$0.27 &   8.93$\pm$0.58 & 00.00$\pm$0.00 & 00.00$\pm$0.00 & 00.00$\pm$0.00 & 00.00$\pm$0.00 & 00.00$\pm$0.00 & 00.00$\pm$0.00 & 00.00$\pm$0.00 & 00.00$\pm$0.00\\
          %& \textit{MF1}   & 87.18$\pm$0.29 &  7.71$\pm$0.78 & 00.00$\pm$0.00 & 00.00$\pm$0.00 & 00.00$\pm$0.00 & 00.00$\pm$0.00 & 00.00$\pm$0.00 & 00.00$\pm$0.00 & 00.00$\pm$0.00 & 00.00$\pm$0.00\\
    %\midrule
    \vspace{2pt}
    \multirow{2}[2]{*} \textit{PatchTST(ICML'22)} & 77.89$\pm$1.98 & 14.42$\pm$0.26 & 68.36$\pm$2.15 & 61.00$\pm$1.36 & 55.00$\pm$2.10 & 90.63$\pm$0.86 & 91.54$\pm$0.41 & 90.25$\pm$0.45 & 59.11$\pm$0.34 & 90.01$\pm$0.30 \\

    \vspace{2pt}
    \multirow{2}[2]{*} \textit{CRT(TNNLS'23)} & 89.21$\pm$0.56 & 7.60$\pm$1.17 & 58.03$\pm$1.11 & 51.67$\pm$2.86 & 54.00$\pm$1.71 & 82.50$\pm$1.17 & 85.81$\pm$0.29 & 91.00$\pm$0.34 & 75.73$\pm$0.08 & 80.87$\pm$0.24\\
          % & \textit{MF1}   & 88.99$\pm$0.40 &  4.94$\pm$1.91 & 54.03$\pm$1.84 & 47.53$\pm$3.34 & 52.30$\pm$1.12 & 82.36$\pm$0.79 & 85.63$\pm$0.12 & 39.50$\pm$2.69 & \underline{76.33$\pm$0.06} & 80.93$\pm$0.17\\
    
    \vspace{2pt}
    \multirow{2}[2]{*} \textit{SimMTM(NeurIPS'23)} & 87.90$\pm$0.35 & 15.45$\pm$0.12 & \textbf{90.78$\pm$0.30} & 62.00$\pm$0.36 & 55.33$\pm$3.21 & 95.42$\pm$0.16 & 95.05$\pm$0.13 & 92.57$\pm$0.18 & 51.98$\pm$0.22 & 91.58$\pm$0.30\\
          % & \textit{MF1}   & 87.94$\pm$0.23 &  \underline{14.60$\pm$0.24} & \textbf{90.78$\pm$0.25} & 61.62$\pm$0.30 & 51.10$\pm$4.82 & 95.84$\pm$0.15 & 95.05$\pm$0.13 & 37.95$\pm$0.07 & 50.48$\pm$0.08 & 91.58$\pm$0.30\\

    \vspace{2pt}
    \multirow{2}[2]{*} \textit{TimeMAE(Arxiv'23)} & 91.31$\pm$0.84 &  14.13$\pm$0.34 & 85.53$\pm$1.84 & \underline{62.60$\pm$1.67} & 53.80$\pm$1.64 & 95.07$\pm$0.72 & 95.76$\pm$0.51 & \underline{93.92$\pm$0.17} & 55.31$\pm$0.33 & 86.78$\pm$1.64\\
          % & \textit{MF1}   & \underline{91.25$\pm$0.06} &   13.12$\pm$0.36 & 85.20$\pm$1.79 & \underline{61.99$\pm$1.32} & \underline{53.67$\pm$1.67} & 95.04$\pm$0.76 & 95.75$\pm$0.51 & \underline{56.71$\pm$0.58} & 54.01$\pm$0.28 & 86.71$\pm$1.65\\

    \midrule
    \multirow{2}[2]{*} \textit{Ours} & \textbf{93.78$\pm$0.70} &  \textbf{18.22$\pm$0.12} & \underline{86.72$\pm$0.98} & \textbf{64.50$\pm$0.20} & \textbf{56.40$\pm$2.94} & \textbf{98.52$\pm$0.29} & \underline{97.98$\pm$0.21} & \textbf{94.51$\pm$0.08} & \textbf{79.57$\pm$0.38} & \textbf{95.99$\pm$0.12}\\
          % & \textit{MF1}   & \textbf{93.32$\pm$0.74} &  \textbf{17.96$\pm$0.36} & \underline{86.62$\pm$1.04} & \textbf{63.30$\pm$0.18} & \textbf{56.06$\pm$3.28} & \textbf{98.57$\pm$0.29} & \underline{97.98$\pm$0.21} & \textbf{60.66$\pm$0.97} & \textbf{79.53$\pm$0.40} & \textbf{95.99$\pm$0.12}\\
    \midrule
    \midrule
    \multirow{2}{*}{} & \multicolumn{10}{c}{\textit{Fine-tune Evaluation}} \\
     & HAR & PS & SRSCP1 & MI & FM & AWR & SAD & ECG5000 & FB & UWare \\
    \midrule

    \vspace{2pt}
    \multirow{2}[2]{*} \textit{TS-TCC(IJCAI'21)} & 91.66$\pm$0.42 &  20.72$\pm$0.68 & 83.76$\pm$1.07 & 57.81$\pm$0.30 & 46.00$\pm$2.12 & 96.61$\pm$0.33 & 98.71$\pm$0.09 & 93.42$\pm$0.61 & 63.68$\pm$0.52 & 89.35$\pm$0.05\\
          % & \textit{MF1}   & 91.86$\pm$0.40 &  19.20$\pm$0.54 & 80.71$\pm$1.03 & 44.57$\pm$0.22 & 45.91$\pm$2.25 & 95.57$\pm$0.41 & 98.71$\pm$0.09 & 55.73$\pm$0.51 & 63.68$\pm$0.41 & 89.31$\pm$0.19\\

    \vspace{2pt}
    \multirow{2}[2]{*} \textit{TS2Vec(AAAI'22)} & 95.10$\pm$0.36 & 16.07$\pm$0.66 & 84.43$\pm$0.61 & 53.00$\pm$0.49 & 49.21$\pm$2.33 & 98.01$\pm$0.06 & 98.36$\pm$0.34 & 93.63$\pm$0.52 & 79.82$\pm$0.14 & 93.45$\pm$0.06\\
          % & \textit{MF1}   & 94.86$\pm$0.42 &   14.63$\pm$1.06 & 84.20$\pm$0.67 & 48.87$\pm$0.48 & 48.51$\pm$2.34 & 98.13$\pm$0.19 & 98.36$\pm$0.30 & 56.84$\pm$0.63 & 79.84$\pm$0.10 & 94.45$\pm$0.03\\
    
    %\multirow{2}[2]{*}{MHCCL} & \textit{Acc.}  & 00.00$\pm$0.00 &  00.00$\pm$0.00 & 00.00$\pm$0.00 & 00.00$\pm$0.00 & 00.00$\pm$0.00 & 00.00$\pm$0.00 & 00.00$\pm$0.00 & 00.00$\pm$0.00 & 00.00$\pm$0.00 & 00.00$\pm$0.00\\
          %& \textit{MF1}   & 00.00$\pm$0.00 &  00.00$\pm$0.00 & 00.00$\pm$0.00 & 00.00$\pm$0.00 & 00.00$\pm$0.00 & 00.00$\pm$0.00 & 00.00$\pm$0.00 & 00.00$\pm$0.00 & 00.00$\pm$0.00 & 00.00$\pm$0.00\\
    %\midrule
    \vspace{2pt}
    \multirow{2}[2]{*} \textit{TS-GAC(AAAI'24)}  & 94.83$\pm$0.10 &  16.76$\pm$0.35 & 86.44$\pm$2.50 & 56.96$\pm$2.54 & 49.11$\pm$4.15 & 98.06$\pm$0.07 & 99.01$\pm$0.20 & - & - & -\\

    \vspace{2pt}
    \multirow{2}[2]{*} \textit{CSL(VLDB'24)} & 89.31$\pm$0.40 & 18.13$\pm$0.48 & 83.95$\pm$0.68 & \underline{64.50$\pm$0.64} & 56.00$\pm$3.27 & \underline{98.66$\pm$0.16} & \textbf{99.44$\pm$0.23} & \underline{94.42$\pm$0.37} & 79.60$\pm$0.30 & \underline{95.53$\pm$0.27} \\
          % & \textit{MF1}   & 94.51$\pm$0.14 &  16.15$\pm$0.29 & 85.67$\pm$2.25 & 55.68$\pm$2.50 & 46.09$\pm$4.57 & 97.94$\pm$0.08 & 98.89$\pm$0.23 & - & - & -\\
    \midrule
    %\multirow{2}[2]{*}{TST} & \textit{Acc.}  & 94.35$\pm$0.11 &  8.82$\pm$0.30 & 00.00$\pm$0.00 & 00.00$\pm$0.00 & 00.00$\pm$0.00 & 00.00$\pm$0.00 & 00.00$\pm$0.00 & 00.00$\pm$0.00 & 00.00$\pm$0.00 & 00.00$\pm$0.00\\
          %& \textit{MF1}   & 94.33$\pm$0.07 &   7.79$\pm$0.91 & 00.00$\pm$0.00 & 00.00$\pm$0.00 & 00.00$\pm$0.00 & 00.00$\pm$0.00 & 00.00$\pm$0.00 & 00.00$\pm$0.00 & 00.00$\pm$0.00 & 00.00$\pm$0.00\\
    %\midrule

    \vspace{2pt}
    \multirow{2}[2]{*} \textit{PatchTST(ICML'22)} & 88.45$\pm$0.70 & 18.42$\pm$0.64 & 81.25$\pm$1.78 & 62.40$\pm$2.35 & 57.81$\pm$2.25 & 98.05$\pm$0.06 & 97.75$\pm$0.16 & 94.40$\pm$0.21 & 75.65$\pm$0.23 & 91.36$\pm$0.17 \\
    
    \vspace{2pt}
    \multirow{2}[2]{*} \textit{CRT(TNNLS'23)}  & 90.09$\pm$0.75 &  8.38$\pm$0.19 & 67.65$\pm$1.36 & 50.67$\pm$2.65 & 53.00$\pm$2.56 & 87.62$\pm$0.84 & 98.23$\pm$0.13 & 92.53$\pm$0.32 & 79.24$\pm$0.13 & 85.64$\pm$0.09\\
          % & \textit{MF1}   & 90.51$\pm$0.77 &  19.04$\pm$0.73 & 67.25$\pm$1.61 & 48.35$\pm$2.89 & 51.69$\pm$2.38 & 87.23$\pm$0.89 & 98.22$\pm$0.14 & 53.62$\pm$0.87 & 79.16$\pm$0.15 & 85.31$\pm$0.16\\

    \vspace{2pt}
    \multirow{2}[2]{*} \textit{SimMTM(NeurIPS'23)} & 93.50$\pm$0.32 &  \underline{21.95$\pm$0.20} & \textbf{92.72$\pm$0.93} & 62.00$\pm$0.20 & 60.00$\pm$2.15 & 98.57$\pm$0.10 & 99.30$\pm$0.10 & 92.76$\pm$0.13 & \textbf{82.53$\pm$0.18} & 94.64$\pm$0.13 \\
          % & \textit{MF1}   & 93.82$\pm$0.10 &  \underline{20.83$\pm$0.31} & \textbf{92.76$\pm$1.20} & \underline{66.67$\pm$0.21} & 59.20$\pm$2.23 & \underline{98.57$\pm$0.10} & \underline{99.30$\pm$0.14} & 40.03$\pm$1.79 & \textbf{82.52$\pm$0.26} & \underline{94.60$\pm$0.10}\\

    \vspace{2pt}
    \multirow{2}[2]{*} \textit{TimeMAE(Arxiv'23)} & \underline{95.11$\pm$0.18} &   19.49$\pm$0.59 & \underline{87.71$\pm$1.27} & 61.80$\pm$2.28 & \textbf{61.00$\pm$1.87} & 97.73$\pm$0.43 &  99.20$\pm$0.03 & 94.25$\pm$0.14 & 67.21$\pm$0.44 & 92.81$\pm$0.47\\
          % & \textit{MF1}   & \underline{95.10$\pm$0.17} &   17.44$\pm$1.15 & \underline{87.96$\pm$1.26} & 61.55$\pm$2.63 & \textbf{64.49$\pm$3.75} & 97.74$\pm$0.42 &  99.20$\pm$0.03 & \underline{57.49$\pm$0.53} & 66.18$\pm$0.50 & 92.66$\pm$0.38\\
    \midrule
    \multirow{2}[2]{*} \textit{Ours} & \textbf{95.92$\pm$0.78} &  \textbf{23.27$\pm$0.37} & 86.25$\pm$1.14 & \textbf{65.20$\pm$0.31} & \underline{60.60$\pm$3.38} & \textbf{98.67$\pm$0.19} & \underline{99.44$\pm$0.12} & \textbf{94.84$\pm$0.10} & \underline{80.83$\pm$0.45} & \textbf{95.85$\pm$0.23}\\
          % & \textit{MF1}   & \textbf{95.91$\pm$0.81} &  \textbf{23.26$\pm$0.34} & 86.18$\pm$1.15 & \textbf{64.65$\pm$0.15} & \underline{59.64$\pm$3.76} & \textbf{98.73$\pm$0.18} & \textbf{99.44$\pm$0.13} & \textbf{60.73$\pm$1.79} & \underline{80.78$\pm$0.45} & \textbf{95.85$\pm$0.23}\\
    \bottomrule
    \bottomrule
    \end{tabular}%
  \caption{Comparisons with State-of-the-Art methods with different evaluations (\%)}
  % \vspace{-2.5em}
  \label{tab:sota}%
  
\end{table*}%

\subsection{Experiment Setup}
\paragraph{Datasets}We conduct experiments on ten publicly available datasets to validate the performance of our model, including Human Activity Recognition (HAR)\cite{anguita2012human} and nine large datasets from the UEA\cite{bagnall2018uea} and UCR archive\cite{dau2019ucr}, which is referred as PS, SRSCP1, MI, FM, AWR, SAD, ECG5000, FB, Uware. These datasets not only encompass both univariate and multivariate datasets, but also cover diverse domains. More details are provided in supplementary materials.

\paragraph{Baselines}We compare our model with recent advanced self-supervised learning methods. To highlight our performance, we select five contrastive learning methods and four masked modeling methods. More details about these baselines are provided in supplementary part:

\begin{itemize}
    \item \textbf{Contrastive Learning Methods: }TS-TCC\cite{eldele2021time}, TS2Vec\cite{yue2022ts2vec}, TS-GAC\cite{wang2024graphaware}, TimesURL\cite{DBLP:conf/aaai/LiuC24}, CSL\cite{DBLP:journals/corr/abs-2305-18888}.
    \item \textbf{Reconstructed Methods: }PatchTST\cite{DBLP:conf/iclr/NieNSK23}, CRT\cite{zhang2023crt}, SimMTM\cite{dong2023simmtm}, TimeMAE\cite{cheng2023timemae}.
\end{itemize}

\paragraph{Evaluation}
In the classification task, we evaluate the performance in two mainstream manner: \textbf{linear evaluation} and \textbf{fine-tuning evaluation}. The former manner freezes the pre-trained parameters of encoder and trains the classifier layer with labeled data. The latter manner updates both the encoder and classifier layer based on the pre-trained parameters. We utilize accuracy as the metrics for classification tasks and mark the \textbf{best} and \underline{second best} values. All experiments are conducted with five different seeds and the average results are taken for comparisons.

\paragraph{Implement Details}
For the encoder, we use the same backbone as TimeMAE with the default 8 transformer layers, while the decoder is configured with 2 layers for both branches. We also adopt the same masking strategy as TimeMAE. We set the batch size as 128 and choose AdamW optimizer with a learning rate of 1e-4. All methods are conducted with NVIDIA A10 and implemented by PyTorch. More implementation and baseline experiments details are provided in supplementary materials.
% In the pre-training stage, we set the layer of transformer-based encoder to 8 following TimeMAE~\cite{cheng2023timemae}. To preserve more information into the encoder and stimulate its encoding capacity, we set the layer of the temporal decoder and frequency decoder to 2. The hidden embedding size in transformer is set to 128 for all datasets. We use AdamW optimizer with a weight decay of 3e-4, $\beta_1 = 0.9$, and $\beta_2 = 0.99$. The learning rate of 1e-4 is adopted. The batch size and mask ratio is set to 128 and 75\% by default, respectively. Due to the distribution difference of frequency information in different datasets, the value of $\gamma$ corresponding to the best performance of each data is slightly different. All methods are run with an NVIDIA A10 and implemented by PyTorch.

\subsection{Quantitative Experimental Results}
Table \ref{tab:sota} presents the classification results. 
Firstly, we observe that our method achieves the highest accuracy across most of datasets, regardless of linear evaluation or fine-tuning task. Particularly, HAR and MI show significant improvements, surpassing the second-best by 2.38\% and 1.90\% in linear probing accuracy. The superior performance of linear evaluation suggests that our method effectively narrows the gap between pretrained representation and task-specific representation, demonstrating strong feature generalization capabilities. Meanwhile, the performance gain on fine-tuning highlight model's adaptability to specific tasks. Notably, we exclude TimesURL from fine-tune evaluation due to its implementation constraints to ensure the rigor of our experiments. 
Secondly, our method supports both univariate and multivariate setting. It is worth noting that TS-GAC considers variables as nodes in graph network, which struggles to deal with single-variable cases. This demonstrates both flexibility and generalization of our method. 
Finally, compared with contrastive-based baselines, our improved masked modeling scheme could strike a balance between dynamic local variation and holistic semantic representation. Compared to masked modeling baselines, our content-aware balanced decoder brings diverse and balanced decoding potential from the spectrum perspective.

\subsection{Analytical experiment}
\begin{figure}[t]
	\centering
	{\includegraphics[width=0.96\linewidth]{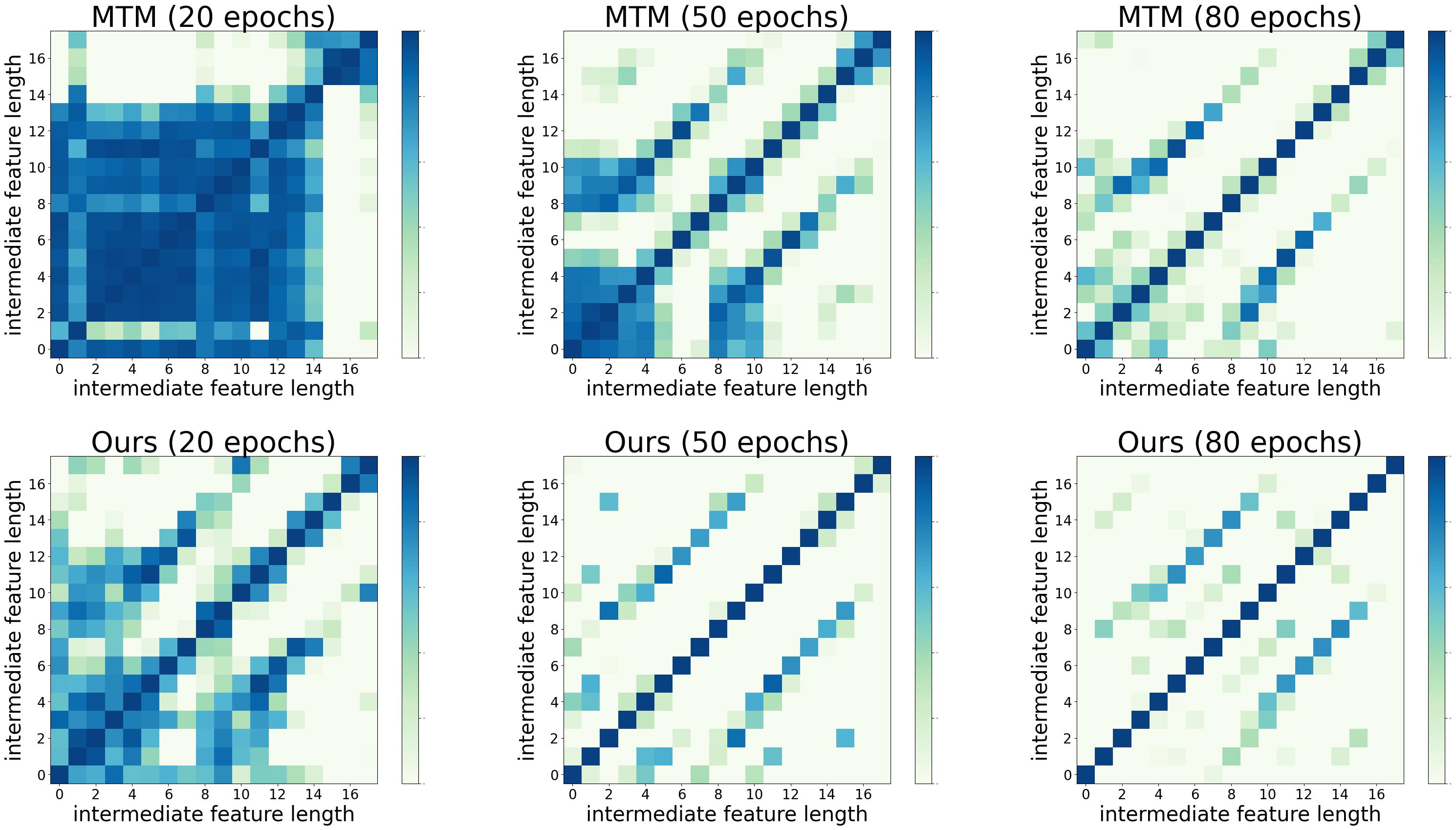}}	
	\caption{Learned Interaction Matrix on the HAR dataset.} 
	\label{fig:view_slope_mat}
\end{figure}

\begin{figure}[t]
	\centering
	{\includegraphics[width=0.96\linewidth]{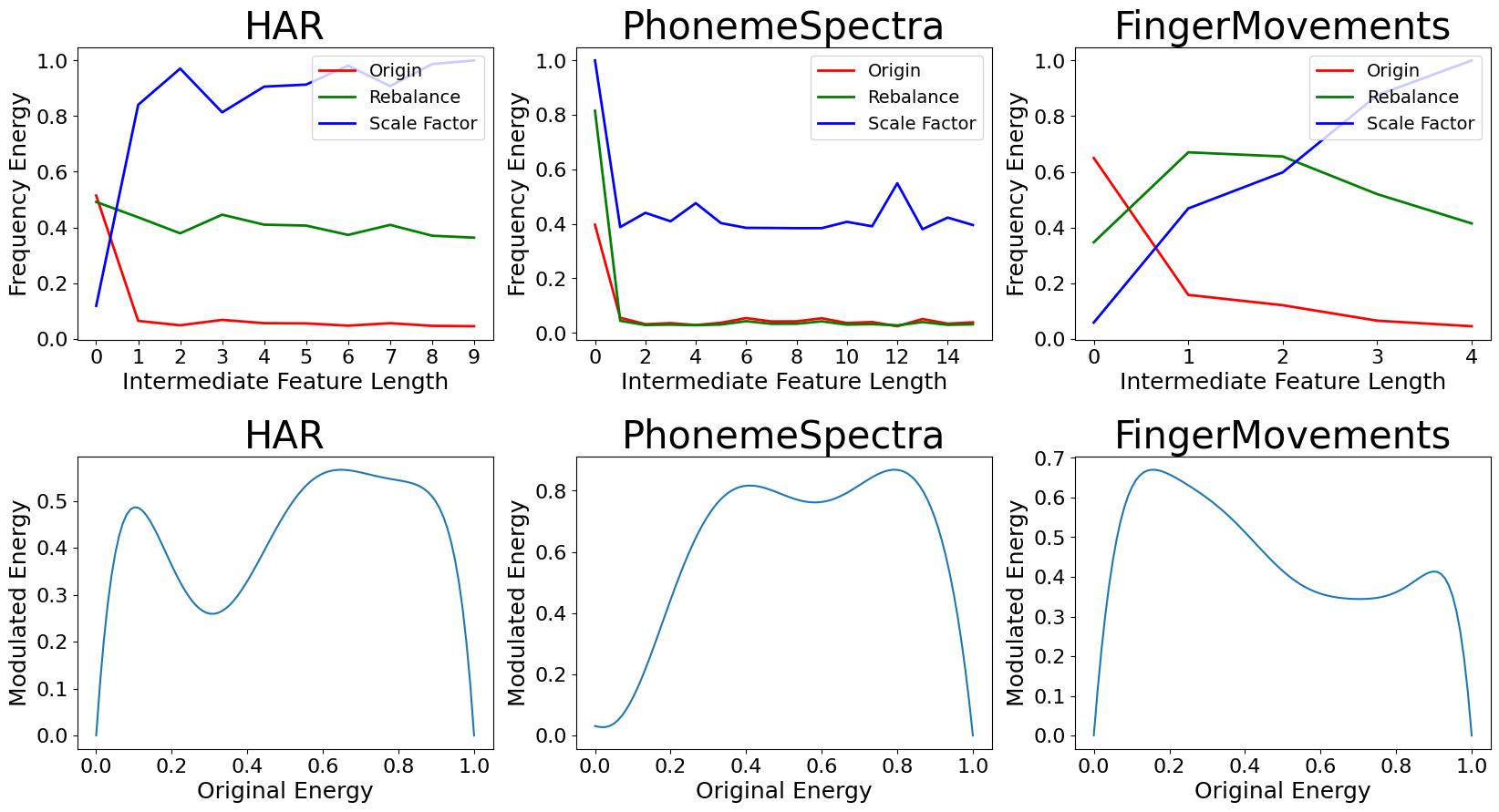}}	
	\caption{Energy rebalance of SER. The top row depicts the modulation during the energy rebalancing, while the bottom represents the corresponding learned Bernstein polynomials.} 
	\label{fig:energy_reb}
\end{figure}

\paragraph{Analysis of Content-aware Interaction Modulation} 
We pretrain our network and vanilla transformer-based reconstruction model on the HAR dataset for 20, 50 and 80 epochs. Then we visualize the interaction matrix obtained from the last encoder layer in Figure~\ref{fig:view_slope_mat}. Both methods exhibit similar encoding patterns in the early stages, emphasizing long-dependency interactions. However, our method employs a more prudent and distinct tactics for interaction scope control in the initial phase. This ability stems from our Content- aware Interaction Modulation Unit (CIM), enabling more accurate and flexible interaction scope compared to the pure self-attention decoding. As training progresses, the encoding receptive fields of both methods gradually shrink, but our approach maintains a more semantically compact receptive field. More visualizations on other datasets are provided in supplementary materials.

\paragraph{Analysis of Spectrum Energy Rebalance}
We investigate an in-depth investigation into the modulation mechanism of the Spectrum Energy Rebalance Unit (SER). Figure~\ref{fig:energy_reb} visualizes the original and modulated spectrum energy distributions for HAR, PhonemeSpectra, and FingerMovements. SER adaptively adjusts the energy distribution by learning data-specific Bernstein coefficients. For HAR (top left), the red line shows the original network concentrating energy in the low-frequency range, indicating a bias towards low-frequency information. As frequency increases, energy decreases, leading to insufficient fitting of mid-to-high frequencies. With SER, the green line shows a marked increase in mid-to-high-frequency energy.
The blue line also illustrates SER's modulation role, which serves as the normalized scaling factor from original energy distribution to balanced one. This flexibility allows SER to compensate for overlooked information and prevent representation degradation. Notably, the model does not excessively prioritize mid-to-high frequencies, as seen in PhonemeSpectra. To further clarify SER's effect, we visualize the learned Bernstein polynomials at the bottom row, summarizing the mapping from normalized energy to the final energy response and highlighting the rebalancing preferences for each dataset.

\paragraph{Ablation Study}
\begin{table}[ht]
  \centering
      % \resizebox{0.7\linewidth}{!}{
    \setlength{\tabcolsep}{10pt}
    \begin{small}
    \begin{tabular}{l|c|c}
    \hline
    \multicolumn{1}{c|}{\multirow{2}{*}{Method}} & LP. & FT. \\
         & Acc.(\%)   & Acc.(\%) \\
    \hline
    \hline
    $\mathcal{L}^{(re)}_{T}$ & 89.47  & 93.34  \\
    $\mathcal{L}^{(re)}_{F}$ & 91.61  & 94.02  \\
    $\mathcal{L}^{(re)}_{T}$ + $\mathcal{L}^{(dual)}_{F}$ & 91.68  & 94.16   \\
    $\mathcal{L}^{(re)}_{F}$ + $\mathcal{L}^{(dual)}_{T}$ & 91.78  & 94.12  \\
    $\mathcal{L}^{(re)}_{T}$ + $\mathcal{L}^{(re)}_{F}$ & 93.37  & 95.61   \\
    \hline
    \hline
    w/o SER & 91.92  & 94.06 \\
    w/o CIM & 93.45  & 95.78  \\
    \hline
    \hline
    \textbf{Ours} & \textbf{93.78} & \textbf{95.92} \\
    \hline
    \end{tabular}%
    \end{small}
  % }
  \caption{Ablation study of designed losses and components in CBD unit.}
  \label{tab:ablation}%
\end{table}%

% K值消融，暂时想着放附录
\begin{table}
	\centering
    \setlength{\tabcolsep}{3pt}
    \begin{small}
	\begin{tabular}{c|cccccc}
		\toprule
        K Value & 1 & 2 & 4 & 8 & \textbf{12} & 16\\
		\midrule
		LP Acc.~(\%) & 92.39 & 92.76 & 93.61 & 93.04 & \textbf{93.78} & 93.48\\
		LP F1~(\%)  & 91.86 & 92.36 & 93.13 & 92.61 & \textbf{93.32} & 93.05\\
		\bottomrule
	\end{tabular}%
     % }
     \end{small}
     \caption{Ablation study of K value in SER.}
    \label{tab:k_value}%
\end{table}
To take a closer look on each part in our framework, we devise several variants and measure their performance using both linear probing (LP) and fine-tuning (FT) on HAR. Results are reported in Table~\ref{tab:ablation} and Table~\ref{tab:k_value}. 
\textbf{(1) Effect of Designed Losses.} Key observations from designed losses include: 
(1) Both the vanilla temporal decoder and our spectrum-aware decoder benefit from the dual-constraint loss. Notably, the frequency-constraint loss $\mathcal{L}_{F}^{(dual)}$, without the frequency decoding branch, improves linear evaluation accuracy by 2.21\%. 
(2) Our frequency decoder outperforms the traditional temporal decoder in feature encoding, achieving improvements of 2.14\% in LP and 0.68\% in FT by addressing feature homogenization and spectrum imbalance. 
(3) The two-pronged decoders could collaboratively improve encoding capability more than either alone, yielding improvements of 3\% and 1.76\% in LP accuracy compared to the standlone temporal and frequency decoders, respectively. Combining them with dual-constraint loss $\mathcal{L}^{(dual)}_{F}$ and $\mathcal{L}^{(dual)}_{T}$ reuslts in the best representation learning performance. 
\textbf{(2) Effectiveness of Each Component in CBD.} Table \ref{tab:ablation} also examines the roles of the CIM and SER units in the frequency decoder. The SER can rebalance the spectrum energy without the CIM. This mechanism can encourage the model to emphasize high-frequency overlooked details. Meanwhile, the CIM dynamically controls the interaction scope based on the local signal variation via flexible large kernel convolutions. With the CIM, the model reduces unnecessary information aggregation and facilitate feature encoding.
\textbf{(3) Performance Variation with Different Order $K$}. To explore the theoretical boundary of Bernstein approximation, we examined performance with varying order $K$ of the Bernstein polynomial in Table~\ref{tab:k_value}. Theoretically, as K approaches infinity, the polynomial can fit any function. However, we found that performance improves with increasing $K$ up to 12, after which it declines. This suggests that $K=12$ is sufficient for SER to learn the scaling factor towards spectrum rebalancing, while higher value will introduce unnecessary noise. Thus, we select $K=12$ as the optimal value.

% TSNE可视化，放附录
% \begin{figure}[ht]
% 	\centering
% 	{\includegraphics[width=0.92\linewidth]{./pic/tsne.jpg}}	
% 	\caption{T-SNE visualization of feature vectors on the HAR dataset, in which each color denotes a specific category} 
% 	\label{fig:tsne}
% \end{figure}

\paragraph{CBD Generality}
\begin{table}[tb]
  \centering
  % \begin{threeparttable}
  \begin{small}
  \setlength{\tabcolsep}{7.5pt}
  \begin{tabular}{c|l|l}
    \toprule
    \scalebox{1.0}{Model} & {\scalebox{1.0}{Avg. LP Acc.}} & {\scalebox{1.0}{Avg. FT Acc.}}\\
    \toprule
    \scalebox{1.0}{PatchTST} & 69.82 & 76.55  \\
    \scalebox{1.0}{\textbf{ + CBD }} & \scalebox{1.0}{\textbf{71.32 (+1.50)}} & \scalebox{1.0}{\textbf{77.56 (+1.01)}} \\
    \midrule
    \scalebox{1.0}{CRT} & \scalebox{1.0}{67.64} & \scalebox{1.0}{71.31} \\
    \scalebox{1.0}{\textbf{ + CBD }} & \scalebox{1.0}{\textbf{72.95 (+5.31)}} & \scalebox{1.0}{\textbf{75.58 (+4.27)}} \\
    \midrule
    \scalebox{1.0}{SimMTM} & \scalebox{1.0}{73.81} & \scalebox{1.0}{79.80} \\
    \scalebox{1.0}{\textbf{ + CBD }} & \scalebox{1.0}{\textbf{74.74 (+0.93)}} & \scalebox{1.0}{\textbf{80.39 (+0.59)}} \\ 
    \bottomrule
  \end{tabular}
  \end{small}
  \caption{Performance by integrating CBD (frequency decoder) to three advanced mask modeling models.}
  \label{tab:rsp_gen}
\end{table}
From Table \ref{tab:rsp_gen}, we can find that the CBD, as an additional decoder, improves the classification performance of diverse advanced masked modeling models. Specifically, for CRT, our module increases the LP accuracy by an average of 5.31\% and FT accuracy by 4.27\% across ten datasets. It is worth mentioning that since TimeMAE does not use the origin data for reconstruction, the CBD may receive inaccurate supervisory signals, leading to performance decline. Detailed experimental results are available in supplementary materials.

\section{Conclusion}
In this paper, we propose a two-pronged reconstruction framework to improve the quality of time-series representation. Specifically, we harness the content-aware balanced decoder with two built-in units to form our powerful frequency decoder, which works collaboratively with the vanilla temporal decoder to effectively address the issues of feature homogenization and spectral energy imbalance. The extensive experiments convincingly demonstrate the superiority of our method based on the observation on quantitative results, behavior analysis, and representation visualization.

\section{Acknowledgments}
This work was supported in part by the National Natural Science Foundation of China, No.:62276155, No.:62376137, No.:U24A20328, No.:62476071, No.:62206156, and No.:62206157; in part by the National Natural Science Foundation of Shandong Province, No.:ZR2021MF040, No.:ZR2024QF104 and No.:ZR2022QF047.

\bibliography{aaai25}

\clearpage

\appendix

\section{A Theorem Proof}

\subsection{A.1 Proof of frequency-domain convolution theorem}
\textbf{Frequency-domain convolution theorem}. The multiplication of two signals in the Fourier domain equals to the Fourier transformation of a convolution of these two signals in their original domain. This can be given by:
\begin{equation}
\fontsize{9pt}{9pt}
\mathcal{F}(\mathbf{K}(v) \otimes \mathbf{Z}(v)) = \mathcal{F}(\mathbf{K}(v)) \odot \mathcal{F}(\mathbf{Z}(v)),
\end{equation}
where $\otimes$ and $\odot$ denote the convolutional operation and element-multiplication operation, and $\mathcal{F}$ refers to the Fourier transformation. $\mathbf{K}(v)$ and $\mathbf{Z}(v)$ represent two signals with respect with time variable $v$.
\newline\textit{Proof}. Suppose that the temporal dimension is denoted as \(T\), then
\begin{equation}
\fontsize{9pt}{9pt}
\mathcal{F}(\mathbf{K}(v) \otimes \mathbf{Z}(v)) = \sum_{i=0}^{T-1} (\mathbf{K}(v_i) \otimes \mathbf{Z}(v_i)) e^{-j 2 \pi f v_i},
\end{equation}
where $j$ represents the imaginary unit. According to convolution theorem, which is written as $\mathbf{K}(v_i) \otimes \mathbf{Z}(v_i) = \sum_{j=0}^{T-1} \mathbf{K}(\tau_j) \mathbf{Z}(v_i-\tau_j)$, then 
\begin{equation}
\fontsize{9pt}{9pt}
\begin{aligned}
\mathcal{F}(\mathbf{K}(v) \otimes \mathbf{Z}(v)) & = \sum_{v = 0}^{T-1} \sum_{\tau =0}^{T-1} (\mathbf{K}(\tau) \mathbf{Z}(v-\tau)) e^{-j 2 \pi f v}\\
& = \sum_{v =0}^{T-1} \sum_{\tau =0}^{T-1} \mathbf{Z}(v-\tau) e^{-j 2 \pi f v} \mathbf{K}(\tau).
\end{aligned}
\end{equation}
Let $x = v - \tau$, then
\begin{equation}
\fontsize{8.5pt}{8.5pt}
\begin{aligned}
\mathcal{F}(\mathbf{K}(v) \otimes \mathbf{Z}(v)) & = \sum_{v = 0}^{T-1} \sum_{\tau = 0}^{T-1} \mathbf{Z}(x) e^{-j 2 \pi f(x+\tau)} x \mathbf{K}(\tau) \\
& = \sum_{v = 0}^{T-1} \sum_{\tau = 0}^{T-1} \mathbf{Z}(x) e^{-j 2 \pi f x} e^{-j 2 \pi f \tau} \mathbf{K}(\tau) \\
& = \sum_{v = 0}^{T-1} \mathbf{K}(\tau) e^{-j 2 \pi f \tau} \sum_{\tau = 0}^{T-1} \mathbf{Z}(x) e^{-j 2 \pi f x} \\
& = \mathcal{F}(\mathbf{K}(v)) \odot \mathcal{F}(\mathbf{Z}(v)).
\end{aligned}
\end{equation}
% This implies that the product of two signals in the Fourier domain is equivalent to the Fourier transformation of the convolution of these two signals in their original domain. 
Proved.

\subsection{A.2 Proof of Parseval’s theorem}
\textbf{Parseval’s theorem.} The energy of feature in temporal domain can be completely characterized by the energy in the frequency domain, which can be formulated as follows:
\begin{equation}
\fontsize{9pt}{9pt}
\sum_{t=-\infty}^{\infty} |\mathbf{Z}(v)|^{2} = \sum_{s=-\infty}^{\infty} |\mathbf{Z}_{F}(\lambda_s)|^{2}, 
\end{equation}
where $\mathbf{Z}$ represents the original temporal domain feature, and $\mathbf{Z}_{F}$ is the corresponding spectrum representation. They satisfy the equation $\mathbf{Z}_{F}(\lambda_s) = \sum_{i=0}^{T} \mathbf{Z}(v_i) e^{-j 2 \pi \lambda_s v_i}$, where $v$ is the temporal variable, $\lambda_s$ denotes the $s$-th frequency component.
\newline\textit{Proof}. Considering the representation of raw time series as $\mathbf{Z} \in \mathbb{R}^{T \times d}$, where $T$ denotes the total length of time series, and we denote the temporal dimension as $v$, then
\begin{equation}
\fontsize{9pt}{9pt}
\sum_{i=0}^{T}|\mathbf{Z}(v_i)|^2 = \sum_{i=0}^{T} \mathbf{Z}(v_i) \mathbf{Z}^*(v_i),
\end{equation}
where $\mathbf{Z}^*(v)$ is the conjugate of $\mathbf{Z}(v)$. According to inverse Fourier transformation, $\mathbf{Z}^*(v_i) = \sum_{s=0}^{S} \mathbf{Z}_{F}^*(\lambda_s) e^{j 2 \pi \lambda_s v_i}$, we can obtain, 
\begin{equation}
\fontsize{9pt}{9pt}
\begin{aligned}
\sum_{i=0}^{T}|\mathbf{Z}(v_i)|^2 & = \sum_{i=0}^{T} \mathbf{Z}(v_i)\left[\sum_{s=0}^{S} \mathbf{Z}_{F}^*(\lambda_s) e^{j 2 \pi \lambda_s v_i} \right] \\
& = \sum_{s=0}^{S} \mathbf{Z}_{F}^*(\lambda_s)\left[\sum_{i=0}^{T} \mathbf{Z}(v_i) e^{j 2 \pi \lambda_s v_i}\right] \\
& = \sum_{s=0}^{S} \mathbf{Z}_{F}^*(\lambda_s) \mathbf{Z}_{F}(\lambda_s)\\
& = \sum_{s=0}^{S}|\mathbf{Z}_{F}(\lambda_s)|^2.
\end{aligned}
\end{equation}
Proved.

\section{B Experiment Setup}

\subsection{B.1 Dataset Details}
We conduct experiments to assess the superiority of our method under linear probing and fine-tuning settings on ten datasets, including Human Activity Recognition (HAR)\cite{anguita2012human} and nine large-scale datasets from the UEA\cite{bagnall2018uea} and UCI\cite{dau2019ucr} archive: PhonemeSpectra (PS), SelfRegulationSCP1 (SRSCP1), MotorImagery (MI), FingerMovements (FM), ArticularyWordRecognition (AWR), SpokenArabicDigits (SAD), ECG5000, FordB and UWaveGestureLibraryAll (UWare). These datasets cover diverse types of signals (human activity recognition, electroencephalography, electrocardiography, speech, sound, motion and sensor), different length (from 50 to 3000) and multivariate channel dimensions (from 1 to 64). When it comes to data processing, for HAR, we directly download the preprocessed files provided by TS-TCC~\cite{eldele2021time} while for the rest we adopt their pre-defined train-test splits. Table~\ref{tab:datasets} summarizes the statistics of each dataset.
\begin{table}
	\centering
    \begin{small}
        \setlength{\tabcolsep}{3pt}
	\begin{tabular}{cccccc}
		\toprule
        Dataset & Example & Length & Channel & Class & Type\\
		\midrule
		HAR & 11,770 & 128 & 9  & 6  & HAR\\
		PS  & 6,668  & 217 & 11 & 39 & SOUND\\
		SRSCP1 & 561  & 896 & 6 & 2 & EEG\\
		MI & 378 & 3,000 & 64 & 2 & EEG\\
		FM & 416 & 50 & 28 & 2 & EEG \\
            AWR & 575 & 144 & 9 & 25 & MOTION \\
            SAD & 8,798 & 93 & 13 & 10 & SPEECH \\
            ECG5000 & 5,000 & 140 & 1 & 5 & ECG \\
            FordB & 4,446 & 500 & 1 & 2 & SENSOR \\
            UWare & 4,478 & 945 & 1 & 8 & HAR \\
		\bottomrule
	\end{tabular}
    \caption{Details of ten widely-used datasets in experiments.}
    \label{tab:datasets}
    \end{small}
\end{table}

\subsection{B.2 Implementation Details}
In the pre-training stage, we set the layer of transformer-based encoder to 8 following TimeMAE~\cite{cheng2023timemae}. To preserve more information into the encoder and stimulate its encoding capacity, we set the layer of the temporal decoder and frequency decoder to 2. The hidden embedding size in transformer is set to 128 for all datasets. We use AdamW optimizer with a weight decay of 3e-4, $\beta_1 = 0.9$, and $\beta_2 = 0.99$. The learning rate of 1e-4 is adopted. The batch size and mask ratio is set to 128 and 75\% by default, respectively. Due to the distribution difference of frequency information in different datasets, the value of $\gamma$ corresponding to the best performance of each data is slightly different. All methods are run with an NVIDIA A10 and implemented by PyTorch~\cite{DBLP:conf/nips/PaszkeGMLBCKLGA19}. 

For the reproduction details on baselines, we adopt the default hyperparameter setting claimed in their paper. For methods that can not provide insufficient hyperparameter details, such as the patch length in CRT~\cite{zhang2023crt}, in their released code repository, we select the optimal one as their final experimental results. 
% This includes configurations such as the config files in SimMTM~\cite{dong2023simmtm} and , etc. 
Notably, CLS~\cite{DBLP:journals/corr/abs-2305-18888} utilizes SVM~\cite{cortes1995support} as the downstream classifier in its code, which differs from the learnable linear classification heads used in most of methods. For fair comparison, we replace it with a linear classification head consistent with TimesURL~\cite{DBLP:conf/aaai/LiuC24} and re-measure the results.

\subsection{B.3 Baselines}
To demonstrate the effectiveness of our plug-and-play frequency decoder in improving various MTM methods, and the superiority of our overall method, we select four advanced MTM frameworks as baselines. Additionally, five contrastive-based approaches are considered to further validate our performance. 
% This comprehensive comparison allows us to assess the advantages of our approach across a diverse range of self-supervised methods in the field.

\begin{enumerate}
    \item Contrastive Learning Approaches:
    \begin{enumerate}
        \item \textbf{TS-TCC}~\cite{eldele2021time} designs a tough cross-view prediction task to perform temporal and contextual contrastive learning.
        \item \textbf{TS2Vec}~\cite{yue2022ts2vec} performs hierarchical contrastive learning to learn multi-scale contextual information at timestamp level and instance level, respectively.
        % which to distinguish different samples. It leverages the principle of contextual consistency to select positive sample pairs.
        \item \textbf{TS-GAC}~\cite{wang2024graphaware} incorporates both node-level and graph-level contrastive strategies into graph learning framework to learn sensor- and global-level features.
        \item \textbf{TimesURL}~\cite{DBLP:conf/aaai/LiuC24} strengthens the universal time-series representations via frequency-temporal augmentation, where double Universums is elaborately constructed as a hard negative.
        \item \textbf{CSL}~\cite{DBLP:journals/corr/abs-2305-18888} utilizes shapelet-based embeddings within a contrastive learning framework to effectively learn generalizable representations for multivariate time series.
    \end{enumerate}
    \item Masked Time-series Modeling Approaches:
    \begin{enumerate}
        \item \textbf{PatchTST}~\cite{DBLP:conf/iclr/NieNSK23} improves long-term forecasting accuracy by segmenting multivariate time series into subseries-level patches, and employing a channel-independent strategy for representation learning.
        \item \textbf{CRT}~\cite{zhang2023crt} proposes a cross-domain dropping-reconstruction task by randomly dropping certain patches in both time and frequency domains to emphasize temporal-spectral correlations.
        \item \textbf{SimMTM}~\cite{dong2023simmtm} reconstructs the original time series by aggregating point-wise representations from multiple masked variations.
        \item \textbf{TimeMAE}~\cite{cheng2023timemae} aligns the reconstructed features with masked features encoded by a momentum encoder and further incorporates discretized encoding to enhance the reconstruction quality.
    \end{enumerate}
\end{enumerate}

% \begin{enumerate}[leftmargin=0pt]
% \item \textbf{TS-TCC}~\cite{eldele2021time} designs a tough cross-view prediction task to perform temporal and contextual contrasting.
% \item \textbf{TS2Vec}~\cite{yue2022ts2vec} performs hierarchical contrastive learning to learn multi-scale contextual information at timestamp level and instance level to distinguish different samples. It leverages the principle of contextual consistency to select positive sample pairs.
% % \item \textbf{MHCCL}~\cite{wilson2020multi} learns unsupervised time series representations via instance-level and cluster-level contrasting based on the contrastive pairs constructed from hierarchical clustering.
% \item \textbf{TS-GAC}~\cite{wang2024graphaware} introduces graph contrasting with both node- and graph- level contrasting to extract sensor- and global-level features.
% % \item \textbf{TST}~\cite{ajakan2014domain} employs a point-wise approach to process time series data and formulates raw time series regression as a self-supervised optimization problem.
% \item \textbf{CRT}~\cite{zhang2023crt} proposes a cross-domain dropping-reconstruction task by randomly dropping certain patches in both time and frequency domains to model temporal-spectral correlations.
% \item \textbf{SimMTM}~\cite{dong2023simmtm} reconstructs the original time series by aggregating point-wise representations from multiple masked variations.
% \item \textbf{TimeMAE}~\cite{cheng2023timemae} applies a mask operation to the preprocessed representations, enhancing information density and improving the efficiency of masking.
% \end{enumerate}

\section{C More Analytical Experiments}

\subsection{C.1 Supplementary Material to Analysis of Content-aware Interaction Modulation}
\begin{figure}[t]
	\centering
	{\includegraphics[width=0.96\linewidth]{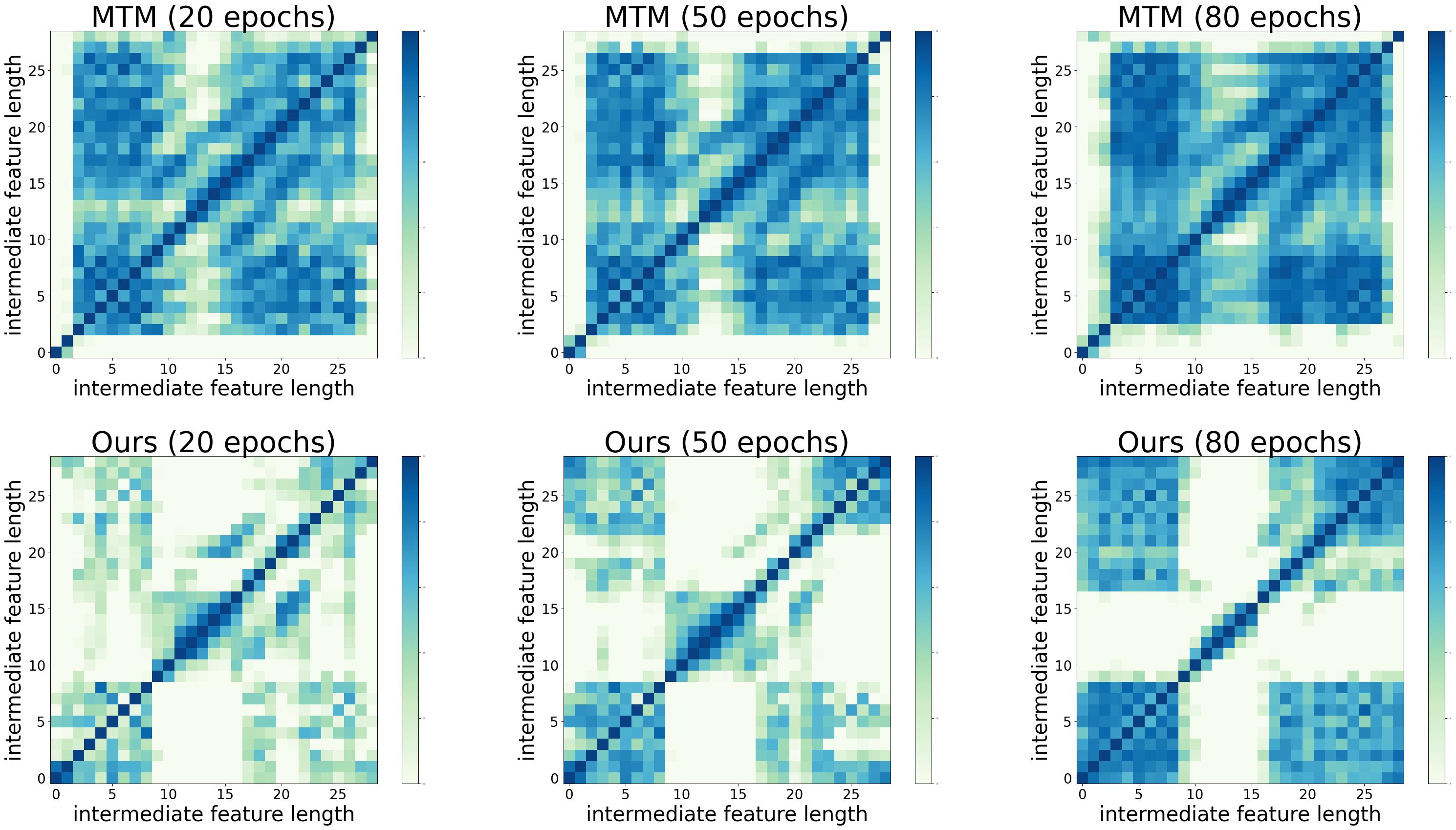}}	
	\caption{Learned Interaction Matrix on the PhonemeSpectra dataset.} 
	\label{fig:view_slope_mat_ps}
\end{figure}

\begin{figure}[t]
	\centering
	{\includegraphics[width=0.96\linewidth]{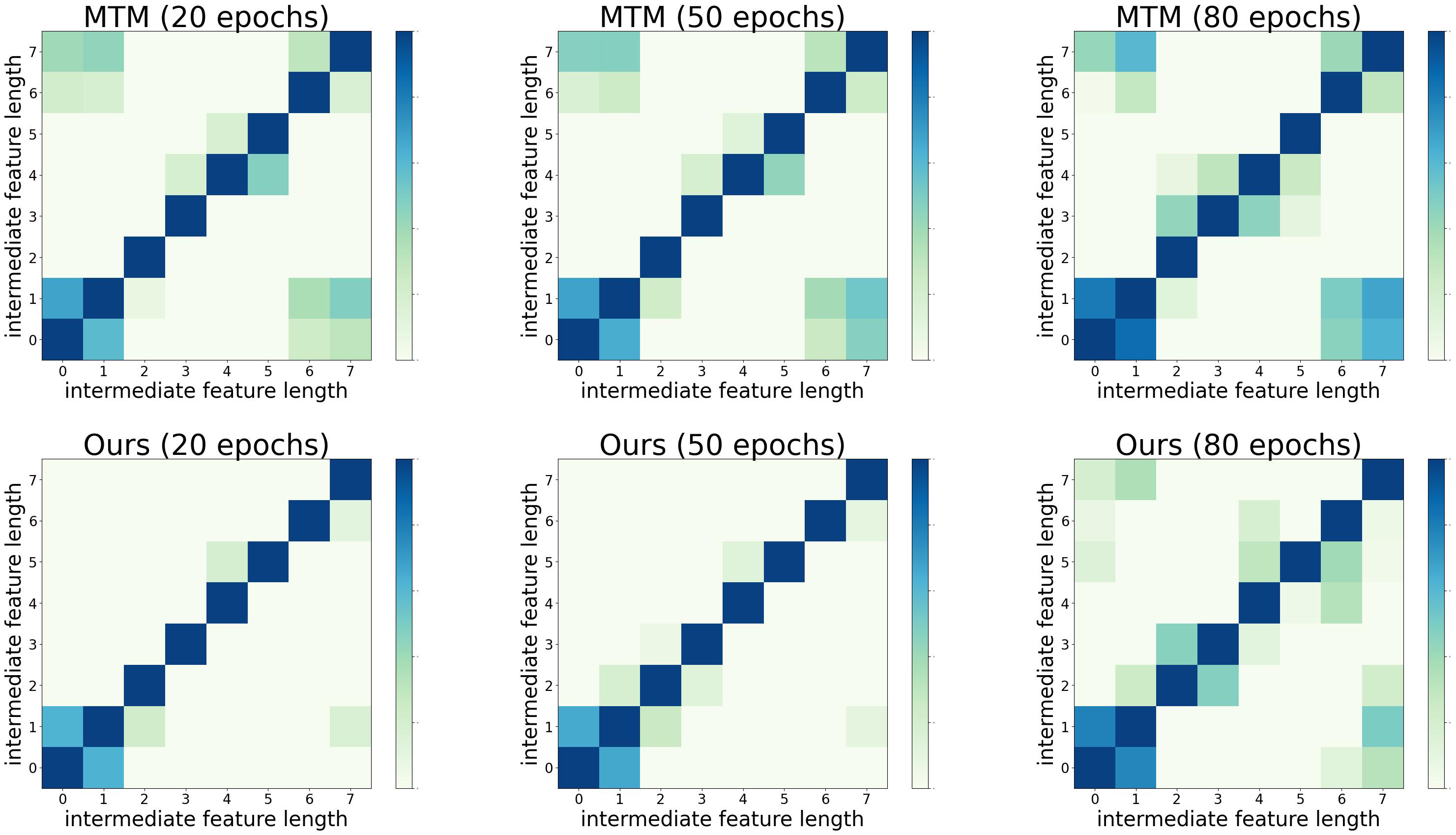}}	
	\caption{Learned Interaction Matrix on the FingerMovements dataset.} 
	\label{fig:view_slope_mat_fm}
\end{figure}

To improve the comprehensiveness of CIM analytic experiment (i.e., only the HAR dataset is provided in the main manuscript), we further include the visualization of interaction matrix regrading PhonemeSpectra and FingerMovements datasets in Figure \ref{fig:view_slope_mat_ps} and Figure \ref{fig:view_slope_mat_fm}. We analyze them from two aspects: (1) we observe that the trend of interactive pattern of these two datasets are different from HAR dataset. On HAR dataset, as training progresses (20 epoch $\rightarrow$ 50epoch $\rightarrow$ 80epoch), the effect of frequency decoder helps the model restrict the scope of receptive interactions. By contrast, the opposite trend regarding PS and FM datasets is exhibited, where the scope of receptive interaction gradually enlarges, supporting delicate expansion of interactive regions. (2) Compared with MTM, the learned interactive matrix of Ours could better mitigate the mindless interactive scope, which effectively optimize the rank of interactive matrix. These visualizations consistently demonstrate the adaptability and flexibility of our Content-aware Interaction Modulation unit in controlling the scope of interaction and mitigating the \textbf{feature homogenization}.

\subsection{C.2 Ablation Study on the Masking Strategy}
In this section, we will provide a detailed description of the masking strategy employed in our approach, along with the ablation study on the masked ratio. 
\paragraph{Masking Strategy} 
\begin{table}
	\centering
    \begin{small}
        \setlength{\tabcolsep}{3pt}
	\begin{tabular}{c|ccc}
		\toprule
        Method & Masked Ratio & Masking Rule & Masking Number\\
		\midrule
		PatchTST & 40\% & Random & 1\\
		CRT & 75\% & Random & 1\\
		SimMTM & 50\% & Random & 3\\
		TimeMAE & 60\% & Random & 1\\
		Ours & 75\% & Random & 1\\
		\bottomrule
	\end{tabular}
    \caption{Masking Strategies of different methods.}
    \label{tab:mask_rule}
    \end{small}
\end{table}
For the masking number, we only randomly the timestamps along temporal dimension once, which is different from SimMTM that masks the timestamps with multiple times (SimMTM does so for better noise reduction). Moreover, we adopt the same masking rules as TimeMAE, where the raw time series is first encoded by a convolutional layer, and the masked to increase the information density of feature. For the masked ratio, we use 75\% masked ratio for all datasets. This value is not the empirical value for these datasets, but it keep pace with the setting of vanilla MTM and CRT. Considering that MTM-based baselines may choose the optimal masked ratio according to their model design, we respect the original works and do not make any additional adjustment to their ratio setting. Instead, we conduct experiments according to the default configurations claimed in their papers, and their masking strategies are elaborated in Table \ref{tab:mask_rule}.

\paragraph{Results with Different Masked Ratio}
\begin{figure}[t]
	\centering
	{\includegraphics[width=0.96\linewidth]{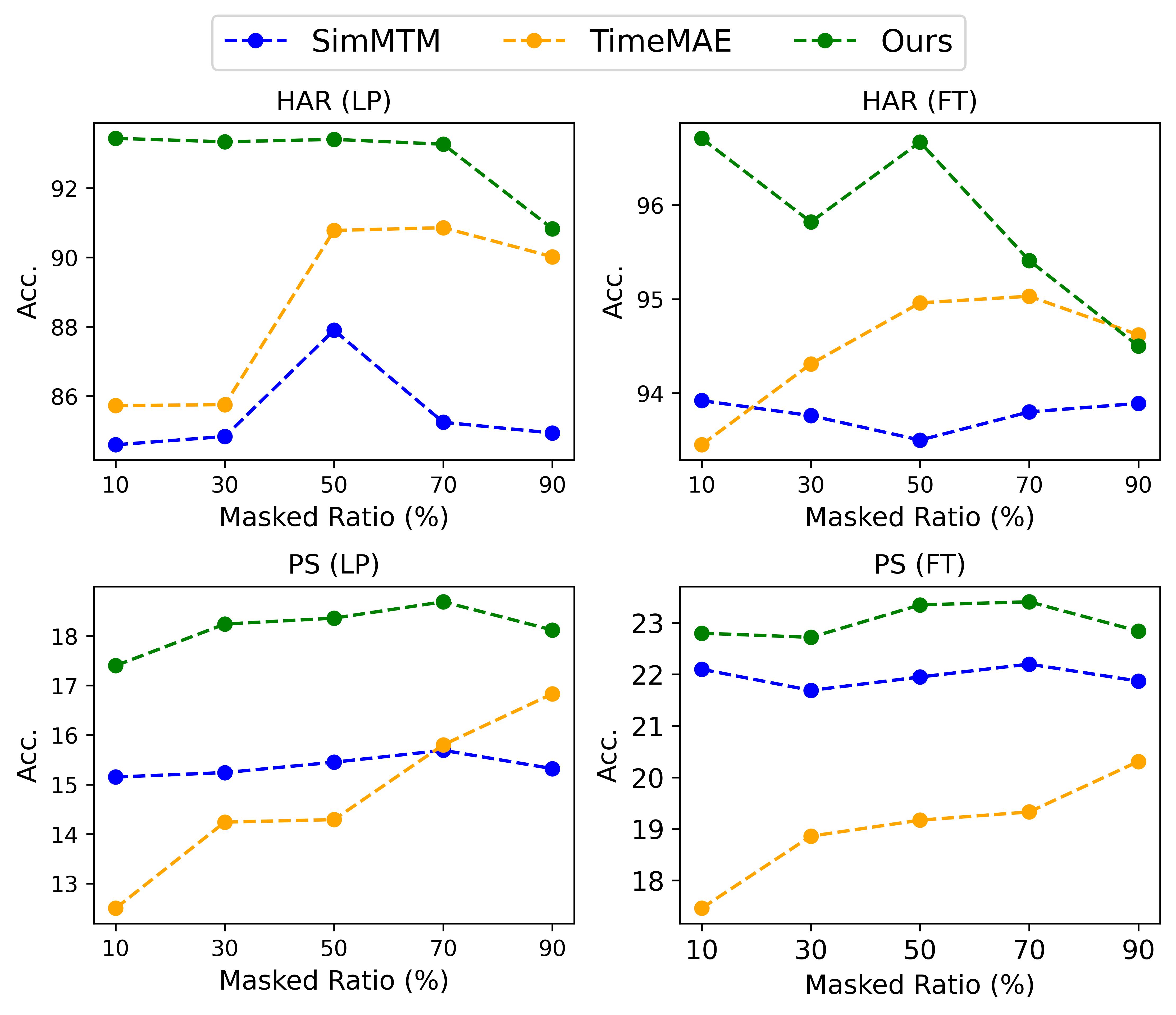}}	
	\caption{Accuracy scores of HAR and PhonemeSpectra of LP and FT with respect to different masked ratios.} 
	\label{fig:mask_exp}
\end{figure}
We report some comparison results regarding mask ratio in Figure\ref{fig:mask_exp}, from which we give a three-point analysis: 
(1) The masked ratio serves as a barometer of the difficulty level of reconstruction task and has a significant impact on the model's representation learning.
(2) The relationship between model performance and masked ratio is indeed not linear. Yet, there is a general trend of an initial increase followed by a decrease in model performance as the masked ratio increases, as shown in MAE\cite{he2022masked} in CV domain, BERT\cite{devlin2018bert} in NLP domain, TimeMAE, SimMTM, and Ours (not limited to these). Specifically, when the masked ratio is low, it fails to stimulate the model to excavate sufficient contextual information thereby learning the satisfactory semantic representation via reconstruction process. As the mask ratio increases, the model learns more contextual information and enhances its generalization ability. However, excessively high mask ratio unavoidably result in the loss of crucial information.
(3) In time-series domain, different datasets exhibit distinct optimal mask ratios. Arguably, for datasets with obvious periodicity, a smaller mask ratio is sufficient for the model to capture the contextual information via reconstruction, because the model could easily memorize the repetitive pattern. By contrast, for datasets with inapparent periodicity, a larger mask ratio allows the model to focus on more comprehensive contextual variation and accurately learn their representation, whereas a smaller mask ratio may be unable to find the complete pattern of time series.

\subsection{C.3 Full Results of CBD Generality}

\begin{table*}[htbp]
  \centering
  \small
  % \tabcolsep=5pt
    % \begin{adjustbox}{width = 1.\textwidth,center}
    % \resizebox{1.0\textwidth}{!}{%
     % \setlength{\tabcolsep}{pt}
    \begin{tabular}{c|cccccccccc}
    \toprule
    \toprule
    \multirow{2}{*}{Methods} & \multicolumn{10}{c}{\textit{Linear Evaluation}} \\
     & HAR & PS & SRSCP1 & MI & FM & AWR & SAD & ECG5000 & FB & UWare \\
    \midrule
    \multirow{2}[2]{*} \textit{PatchTST} & 77.89 &  14.42 & 68.36 & 61.00 & 55.00 & 90.63 & \textbf{91.54} & \textbf{90.25} & 59.11 & 90.01 \\

    \multirow{2}[2]{*} \textit{PatchTST + CBD} & \textbf{78.84} & \textbf{15.96} & \textbf{75.78} & \textbf{61.20} & \textbf{59.37} & \textbf{92.97} & 89.20 & 90.20 & \textbf{59.51} & \textbf{90.13}\\

    \midrule

    \multirow{2}[2]{*} \textit{CRT}  & 89.21 & 7.60 & 58.03 & 51.67 & 54.00 & 82.50 & 85.81 & 91.00 & 75.73 & 80.87 \\

    \multirow{2}[2]{*} \textit{CRT + CBD} & \textbf{92.32} & \textbf{9.16} & \textbf{71.00} & \textbf{54.12} & \textbf{56.91} & \textbf{88.75} & \textbf{88.94} & \textbf{93.01} & \textbf{92.89} & \textbf{82.43} \\

    \midrule

    \multirow{2}[2]{*} \textit{SimMTM} & 87.90 & 15.45 & 90.78 & 62.00 & \textbf{55.33} & 95.42 & 95.05 & 92.57 & \textbf{51.98} & 91.58\\
    
    \multirow{2}[2]{*} \textit{SimMTM + CBD}  & \textbf{88.12} & \textbf{17.54} & \textbf{90.79} & \textbf{63.00} & 55.30 & \textbf{96.83} & \textbf{97.68} & \textbf{93.35} & 51.92 & \textbf{92.88} \\
    \midrule
    \midrule
    \multirow{2}{*}{} & \multicolumn{10}{c}{\textit{Fine-tune Evaluation}} \\
     & HAR & PS & SRSCP1 & MI & FM & AWR & SAD & ECG5000 & FB & UWare \\
    \midrule

    \multirow{2}[2]{*} \textit{PatchTST} & \textbf{88.45} &  18.42 & 81.25 & 62.40 & 57.81 & 98.05 & 97.75 & 94.40 & 75.65 & 91.36\\

    \multirow{2}[2]{*} \textit{PatchTST + CBD} & 87.81 & \textbf{19.35} & \textbf{84.77} & \textbf{63.00} & \textbf{59.38} & \textbf{98.44} & \textbf{97.79} & \textbf{94.44} & \textbf{79.17} & \textbf{91.48}\\

    \midrule

    \multirow{2}[2]{*} \textit{CRT}  & 90.09 &  8.38 & 67.65 & 50.67 & 53.00 & 87.62 & 98.23 & 92.53 & 79.24 & 85.64\\

    \multirow{2}[2]{*} \textit{CRT + CBD} & \textbf{92.39} & \textbf{9.61} & \textbf{73.78} & \textbf{55.11} & \textbf{55.57} & \textbf{93.73} & 98.23 & \textbf{94.91} & \textbf{93.19} & \textbf{89.32} \\

    \midrule

    \multirow{2}[2]{*} \textit{SimMTM} & 93.50 & 21.95 & 92.72 & 62.00 & 60.00 & \textbf{98.57} & 99.30 & 92.76 & \textbf{82.53} & 94.64 \\
    
    \multirow{2}[2]{*} \textit{SimMTM + CBD}  & \textbf{94.27} & \textbf{24.37} & \textbf{92.83} & \textbf{65.00} & \textbf{62.00} & 98.53 & \textbf{99.54} & \textbf{94.04} & 80.37 & \textbf{95.90} \\

    \bottomrule
    \bottomrule
    \end{tabular}%
    % }
  \caption{Full evaluation results of the CBD scalability and generality based on other MTM Methods (\%)}
  % \vspace{-2.5em}
  \label{tab:full_gen}%
\end{table*}%

Table \ref{tab:full_gen} presents the full results of our Content-aware Balanced Decoder (CBD) with other Masked Time-series Modeling Methods. Specifically, we maintain the original reconstruction branch (i.e., temporal decoder) and add a frequency reconstruction branch (i.e., CBD). 
% The frequency information reconstructed by CBD is then subjected to an inverse Fourier transform to compute the reconstruction loss as utilized in the original work. 
We set the layer number of CBD to 2, and set the balanced weight $\gamma$ to 0.5.
We can observe that across most datasets, our CBD significantly enhances the classification performance of the original MTM architecture. This indicates that CBD provides more diverse and balanced feature benefited from the spectrum space.

\subsection{C.4 Visualization Analysis}
\begin{figure}[t]
	\centering
	{\includegraphics[width=0.96\linewidth]{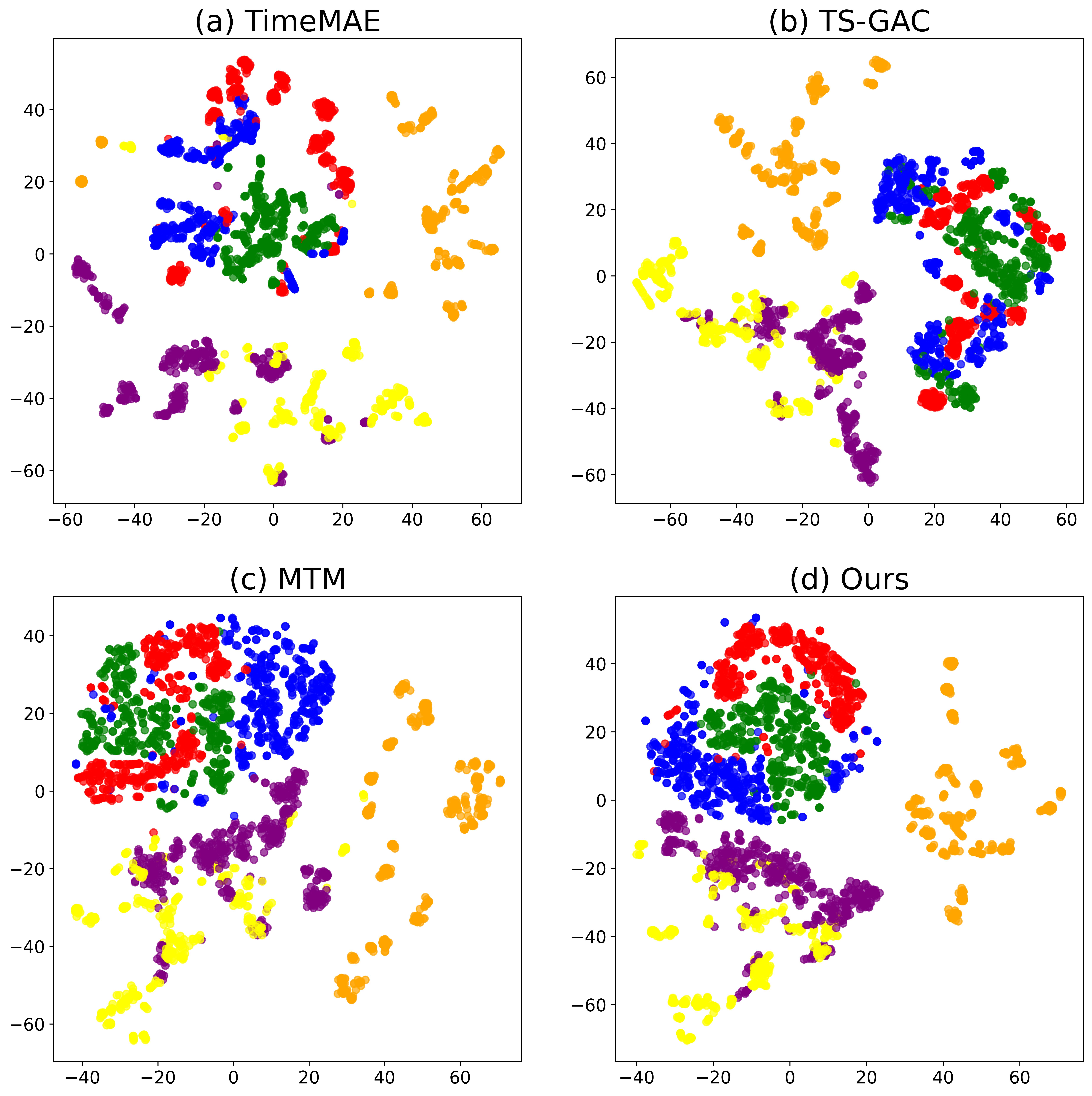}}	
	\caption{T-SNE visualization of feature vectors on the HAR dataset.} 
	\label{fig:tsne_har}
\end{figure}
\begin{figure}[t]
	\centering
	{\includegraphics[width=0.96\linewidth]{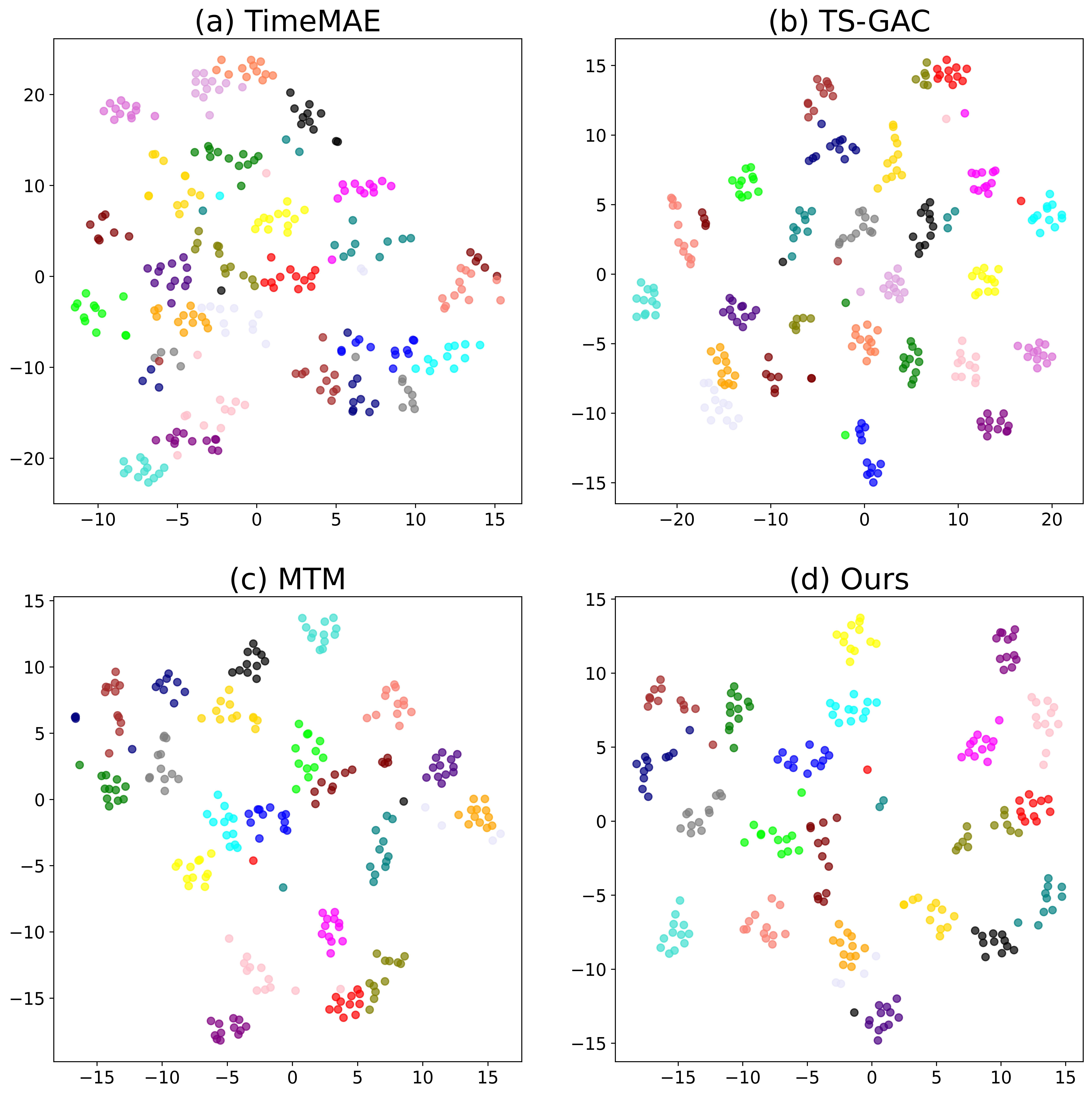}}	
	\caption{T-SNE visualization of feature vectors on the ArticularyWordRecognition dataset.} 
	\label{fig:tsne_awr}
\end{figure}
To demonstrate the superiority of our method on representation learning, we employ T-SNE~\cite{van2008visualizing} algorithm to display the features learned from HAR and ArticularyWordRecognition dataset on several methods in Figure~\ref{fig:tsne_har} and Figure~\ref{fig:tsne_awr}. Specifically, (a) and (b) show the representations obtained by TimeMAE and TS-GAC, respectively. (c) shows the vanilla masked time-series model, and (d) illustrates the visualized results from our overall model. We observe that our method exhibits better class boundary separability compared to the other three methods, particularly on the classes marked by color red, green and blue in HAR. This observation underlines the promising role of our overall framework in enhancing representation discriminative learning.

\end{document}